\documentclass{clv3}

\usepackage{multirow}
\usepackage{hyperref}
\usepackage{ragged2e}
\usepackage{pbox}
\usepackage{arydshln}
\usepackage{graphicx}
\usepackage[]{mdframed}
\definecolor{darkblue}{rgb}{0, 0, 0.5}
\hypersetup{colorlinks=true,citecolor=darkblue, linkcolor=darkblue, urlcolor=darkblue}
\usepackage{svg}
\usepackage{tikz}
\usepackage{epstopdf}
\usepackage[utf8]{inputenc}
\usepackage{graphicx}
\usepackage{hyperref}
\usepackage{graphicx}
\usepackage{tabularray}
\usepackage{xcolor}

\usepackage{color}
\usepackage{algorithm}
\usepackage{algorithmicx}
\usepackage{algpseudocode}%
\usepackage{listings}%
\usepackage{pbox}
\usepackage{arydshln}
\usepackage{booktabs}
\usepackage{subfig}
\usepackage{longtable}
\usepackage[utf8]{inputenc}

\newcounter{para}

\begin{document}
\issue{1}{1}{2024}

\dochead{BhashaVerse : Translation Ecosystem for Indian Subcontinent Languages}

\runningtitle{BhashaVerse : Translation Ecosystem for Indian Subcontinent Languages}

\runningauthor{Vandan Mujadia}

\author{Vandan Mujadia\thanks{Corresponding author}}
\affil{Language Technologies Research Centre,\\ IIIT-Hyderabad, Hyderabad, \\Telangana, India\\vmujadia@gmail.com}

\author{Dipti Misra Sharma}
\affil{Language Technologies Research Centre,\\ IIIT-Hyderabad, Hyderabad, \\Telangana, India\\diptims@gmail.com}

\maketitle

\begin{abstract}
We present a multilingual, multi-task translation model for 36 Indian languages, including Assamese, Hindi, Tamil, and Urdu. The model utilizes 10 billion parallel corpora for 36 x 36 language directions and millions of data pairs for tasks like translation evaluation, post-editing, grammar correction, and error identification. Corpus creation involves leveraging existing resources, synthetic data, and domain-specific datasets. The model’s multi-task approach integrates translation, grammar correction, error identification, quality estimation, and post-editing, ensuring robust cross-lingual communication. Evaluations cover discourse-level and domain-specific translation, using reference-based and reference-free metrics. This work enhances translation quality, supports linguistic diversity, and advances low-resource language processing.
\end{abstract}

\section{Introduction}
Generative Large Language Models (LLMs) have brought transformative advancements to natural language processing (NLP), excelling in a wide range of applications \cite{xuanfan-piji-2023-systematic, xi2023rise}. These models demonstrate remarkable capabilities in open-domain question answering by generating accurate and coherent responses and performing instruction-based tasks such as code completion, error detection, and correction \cite{vaithilingam2022expectation}. Additionally, LLMs are highly effective in tasks like essay writing, grammar correction \cite{wu2023chatgpt}, and text summarization, producing outputs of exceptional quality \cite{chang2023survey}. However, much of this progress has been concentrated on English, leaving many other languages underexplored or underdeveloped \cite{lai2023chatgpt, zhu2023multilingual}. For machine translation (MT) and related subtasks, encoder-decoder-based models remain state-of-the-art, while decoder-only models require further advancement to effectively handle low-resource languages, particularly those spoken in India.  \\

India's linguistic diversity is immense, comprising 22 scheduled languages from linguistic families such as Indo-Aryan, Dravidian, Tibeto-Burman, and Austroasiatic. There are over 559 classified mother tongues spoken across the country, reflecting the cultural and linguistic richness of the nation\footnote{\url{https://en.wikipedia.org/wiki/Linguistic_Survey_of_India}}. This diversity, however, poses significant challenges in communication, education, business, healthcare, tourism, and governance. Advances in encoder-decoder and LLM-based models hold the potential to address these linguistic challenges by enabling cross-lingual communication and fostering India's multilingual ecosystem.\\

Translation plays a vital role in addressing linguistic diversity and enabling effective cross-lingual communication in multilingual societies like India. However, the distinct characteristics of Indian languages present unique challenges. Many Indian languages exhibit complex morphological structures, often involving agglutinative or inflectional morphology, which makes them syntactically diverse and distinct from each other and English. The use of diverse scripts, such as Devanagari, Tamil, Bengali, and Gurmukhi, adds further complexity to text processing tasks. Additionally, code-mixing, a common phenomenon in India where speakers frequently blend multiple languages in communication, complicates translation workflows. The scarcity of high-quality linguistic resources, annotated corpora, and evaluation benchmarks for many Indian languages exacerbates these challenges, hindering the development of robust MT systems.\\

Addressing these challenges requires a systematic approach. Developing translation systems for Indian languages is crucial to facilitate communication and preserve linguistic diversity. With the translation system, robust evaluation frameworks are needed to assess their quality, with or without reference translations. Identifying and categorizing translation errors is essential for system improvement, enabling targeted refinements based on predefined error groups. Automatic post-editing systems can then help correct translation errors efficiently, either independently or in collaboration with human translators. Incorporating these advancements across general and domain-specific applications will greatly benefit the language research community and end users.\\

This paper focuses on developing translation models and related applications for 36 Indian languages, including Assamese, Awadhi, Bengali, Bhojpuri, Braj, Bodo, Dogri, English, Konkani, Gondi, Gujarati, Hindi, Hinglish, Ho, Kannada, Kangri, Kashmiri (Arabic and Devanagari), Khasi, Mizo, Magahi, Maithili, Malayalam, Marathi, Manipuri (Bengali and Meitei), Nepali, Oriya, Punjabi, Sanskrit, Santali, Sinhala, Sindhi (Arabic and Devanagari), Tamil, Tulu, Telugu, and Urdu. Achieving this requires parallel and other types of corpora for all 36 × 36 language pairs, addressing challenges like script variations, phonetic differences, and syntactic diversity. For instance, languages like Kashmiri and Sindhi, which use multiple scripts, demand script normalization for alignment, while low-resource languages such as Khasi and Santali require synthetic data augmentation to ensure sufficient coverage and quality.\\

To address these challenges, this work proposes strategies for corpus creation by leveraging existing resources, developing parallel datasets, generating domain-specific corpora, and utilizing synthetic data techniques. Additionally, it evaluates machine translation across various dimensions, including standard and discourse-level translation, domain-specific translation, reference-based and reference-free evaluation, error analysis, and automatic post-editing. By integrating these elements, the study establishes a comprehensive framework to improve machine translation quality and enable better cross-lingual communication in India's linguistically diverse ecosystem.\\

Finally, this work addresses the following translation-related tasks that are central to enhancing multilingual communication in the Indian subcontinent.

\begin{enumerate}
\justifying
\item \textbf{Translation Across 36 × 36 Indian Language Pairs:}  We develop machine translation models capable of translation for 36 Indian languages to 36 Indian languages. This involves creating n-way parallel corpora, aligning language pairs, and optimizing translation quality to meet linguistic and cultural expectations.  

\item \textbf{Discourse Translation:} Translation at the discourse level addresses coherence, context retention, and inter-sentence dependencies. Our work evaluates how well current systems handle these aspects and proposes improvements for translating connected text effectively. Discourse translation possible for 36 x 36 language pairs with keeping 1536 tokens as a context with encoder-decoder model while 4096 context for decoder only model.

\item \textbf{Domain-Specific Translation:} Different domains, such as healthcare, legal, and education, require specialized vocabulary and context-aware translations. We design systems capable of handling domain-specific language nuances to improve usability in practical scenarios. 

\item \textbf{Machine Translation Evaluation:}  We introduce both reference-based and reference-free evaluation methods to assess translation quality. While reference-based methods compare MT output against human-translated gold standards, reference-free approaches evaluate fluency, adequacy, and fidelity without explicit references.  

\item \textbf{Translation Error Identification and Categorization:}  To identify and address translation errors, we propose a taxonomy for error analysis that includes lexical, grammatical, semantic, and discourse-level errors. This categorization aids in pinpointing system weaknesses and guiding future improvements.  

\item \textbf{Automatic Post-Editing (APE):} Post-editing involves correcting errors in machine-generated translations to improve fluency and accuracy. We develop automatic post-editing systems tailored for Indian languages, leveraging error patterns to refine translations.  

\end{enumerate}

By addressing these interconnected challenges, our work aims to advance the state of machine translation for Indian languages, bridging linguistic gaps and fostering multilingual communication in diverse settings. The remainder of this paper is organized as follows: Section 2 presents related work in machine translation and evaluation for low-resource languages, with a focus on Indian languages. Section 3 details the data collection and preprocessing methods used for building parallel corpora and specialized datasets. Section 4 describes the translation models and architectures employed for sentence, discourse, and domain translation tasks. Section 5 elaborates on the evaluation framework, including metrics and human assessment strategies. Section 6 introduces our taxonomy for error identification and discusses the implementation of automatic post-editing systems. Section 7 presents experimental results, analysis, and comparisons with existing systems. Section 8 concludes the paper with a discussion of findings and future research directions.

\section{Corpora : Existing Machine Translation Corpora}

Corpora are essential for developing language processing systems, especially for low-resource, morphologically complex languages like those in the Indian subcontinent. They underpin the training, evaluation, and tuning of machine translation (MT) models, ensuring diverse and accurate outputs. This section explores existing and newly developed corpora for MT, focusing on preparation, cleaning, alignment, and generation. Indian languages pose unique corpus creation challenges due to script, grammar, and cultural diversity. We use a mix of curated parallel corpora, domain-specific datasets, and synthetic data to address these, facilitating robust translation models for 36 Indian languages and their combinations.\\

Machine translation (MT) research for Indian languages is heavily reliant on robust parallel corpora that address the unique linguistic, resource, and domain-specific challenges inherent in India’s multilingual landscape. Over the years, a variety of datasets have been developed to support translation between Indian and other languages, as well as Indian-to-Indian language pairs, enabling significant progress in this field.\\

The BPCC Parallel Corpus \citep{gala2023indictrans2} represents the largest publicly accessible resource, a collection of several existing and developed parallel corpora resources, consisting of 230 million sentence pairs across English and 22 Indian languages. It features the BPCC-Human Corpus, a refined subset containing 2.2 million English-Indic sentence pairs that have been manually validated, thereby providing dependable data for both training and evaluation purposes. In a similar context, the Samanantar Parallel Corpus \citep{ramesh2022samanantar} was, until 2021, the most extensive collection of parallel corpora for English and 11 Indian languages, offering 46 million English-Indic sentence pairs and 82 million Indian-to-Indian language pairs. This corpus serves as a crucial dataset for multilingual machine translation research.\\

Corpora such as the IIT Bombay English-Hindi Parallel Corpus (1.5 million segments) \citep{kunchukuttan2017iit} and PMIndia \citep{haddow2020pmindia} provide domain-specific resources, particularly for English-Hindi and other English-Indic translations. PMIndia, derived from government texts like the Prime Minister’s Mann Ki Baat speeches \citep{philip2019cvit}, and the CVIT-IIITH PIB Multilingual Corpus \citep{philip2019cvit}, mined from Press Information Bureau\footnote{\url{https://pib.gov.in/}} web pages, emphasize high-quality, formal translations. These datasets enrich MT research with domain-specific data critical for practical applications.\\

For Indian-to-Indian language translation, resources are more limited. The LTRC Parallel Corpus\footnote{\url{https://github.com/vmujadia/The-LTRC-Hindi-Telugu-Parallel-Corpus}} for Hindi-Telugu \citep{mujadia-sharma-2022-ltrc}, developed by IIIT Hyderabad, offers 500,000 sentence-aligned translations, providing robust data for these structurally distinct languages. Another work focusing on a low-resource language pair, Hindi-Kangri, presented a corpus \citep{chauhan2021monolingualparallelcorporakangri} with 27,000 parallel sentences. The IndoWordNet Parallel Corpus \citep{bhattacharyya2020indowordnet}, with 6.3 million segments across 18 languages, is another key resource, leveraging glosses and examples to facilitate Indian-to-Indian language MT development.\\

Low-resource languages are addressed through specialized corpora such as the Ema-lon Manipuri Corpus (124,975 Manipuri-English aligned sentences) \citep{huidrom2021corpus} and NLLB-Seed\footnote{\url{https://github.com/facebookresearch/flores/blob/main/nllb_seed/README.md}, \url{https://github.com/openlanguagedata/seed}}, which focuses on low resourced languages like Kashmiri, Maithili, and Bhojpuri among others. These efforts aim to provide critical data for languages often underrepresented in MT research. Code-mixed languages, a common phenomenon in India, are supported by datasets like PHINC \citep{srivastava2020phinc} (Hinglish-English, 13,738 sentences) and the IIIT-H en-hi-codemixed-corpus \citep{dhar2018enabling} (6,096 English-Hindi code-mixed sentences), which are essential for handling multilingual and code-switched input.\\

Domain-specific corpora further enhance the scope of Indian languages machine translation, Itihasa Parallel Corpus \citep{aralikatte2021itihasa} offers 93,000 parallel sentences for English-Sanskrit translations, catering to ancient language research, while the TED Parallel Corpus\footnote{\url{https://github.com/ajinkyakulkarni14/TED-Multilingual-Parallel-Corpus}} and JW300 Corpus \citep{agic2019jw300} provide resources for public speeches and religious texts. Regional languages are supported by corpora like CGNetSwara\footnote{\url{http://cgnetswara.org/}} Hindi-Gondi (19,000 sentence pairs) and MTEnglish2Odia\footnote{\url{https://github.com/soumendrak/MTEnglish2Odia}} (42,000 pairs), filling critical gaps for underrepresented groups.\\

Large-scale mining initiatives like CCAligned (100 million document pairs across 137 languages)\citep{el2019ccaligned} and CoPara \citep{nikhil2023copara}(passage-level alignments for Dravidian languages) significantly expand multilingual datasets. While noisy, resources like WikiMatrix\citep{schwenk2019wikimatrix} and CCMatrix \citep{schwenk2019ccmatrix} mined from Wikipedia and CommonCrawl offer valuable data for pre-processing pipelines. Other notable resources include the Nepali National Corpus \citep{yadava2008construction}, the Kathmandu University English-Nepali Parallel Corpus \citep{duwal2019efforts}, and the Uka Tarsadia University Corpus for English-Gujarati translations \citep{Shah_2019}, each addressing specific regional and linguistic needs.\\

The Low-Resource MT Shared Task \footnote{\url{https://sites.google.com/view/loresmt/}} \footnote{\url{https://www.statmt.org/wmt21/similar.html}} \footnote{\url{https://github.com/loresmt}} Corpus, used in platforms like the Workshop on Asian Translation (WAT) \footnote{\url{https://lotus.kuee.kyoto-u.ac.jp/WAT/WAT2024/index.html}}, supports MT systems for Hindi and other low-resource languages. These datasets combine Wikipedia, government documents, and religious texts with crowd-sourced translations, focusing on language pairs like Hindi-Marathi, Hindi-Telugu, Tamil-Telugu, and Hindi-Bhojpuri. Synthetic techniques like back-translation \citep{gala2023indictrans2} and multilingual pretraining are increasingly used to augment these corpora, improving model performance for low-resource languages. These efforts drive innovation in transfer learning, multilingual modeling, and domain-specific translation, addressing critical gaps in India’s MT landscape.\\

Collectively, these corpora form a robust foundation for machine translation research and development work for Indian languages, advancing cross-lingual communication and supporting the preservation of India’s linguistic diversity. 

\section{Developed Corpora : Machine Translation}

In this section, we detail the methodologies that we used to build corpora for training machine translation (MT) systems, focusing on Indian subcontinent languages. We explore alignment techniques, domain-specific corpora development, synthetic data generation methods, and corpus cleaning methodologies to build more than 1B parallel corpora for machine translation involving 36 Indian subcontinent languages.

\subsection*{Automatic Alignment and Human Validation}

We collected data from various multilingual websites and books containing content in English and multiple Indian languages. Using a neural alignment tool\footnote{\url{https://github.com/vmujadia/sentencealigner}} based on the COMET-QE model \citep{rei-etal-2022-comet}, we calculated sentence-level similarity scores to align multilingual webpage content. This allowed us to identify English sentences and their corresponding translations in other languages, facilitating the creation of parallel corpora. For each language pair, specific alignment thresholds\footnote{determined based on the average COMET-QE score for the respective language pair} were applied, enabling the extraction of aligned data for pairs such as English-Assamese, English-Bangla, English-Gujarati, English-Hindi, English-Kannada, English-Malayalam, English-Marathi, English-Odia, English-Punjabi, English-Tamil, and English-Telugu. \\

The process presented challenges due to the diverse sentence structures, significant word order variations, and rich morphological features inherent to Indian languages. To address these, we conducted human validation on 10\% of the sampled data to assess and adjust the alignment thresholds, ensuring the creation of high-quality parallel corpora suitable for training machine translation systems.

\subsection*{Human Post-edited Education and Medical Domain Parallel Corpora}
\label{sec:posteditedcorpora}
Domain-specific corpora are essential for developing machine translation (MT) systems that perform exceptionally well in specialized areas such as medicine and technical content. These corpora capture critical domain-specific terminology, language expressions, and contextual nuances, disfluencies managed in translation if the content is spoken, which are often missing from general-purpose datasets. \\

To build parallel corpora, we enlisted multiple freelance translation experts to translate text from English into Indian languages, including Assamese, Gujarati, Hindi, Kannada, Malayalam, Marathi, Odia, Tamil, and Telugu. The primary focus was on educational domains, utilizing text collected from technical lectures, including those sourced from the Swayam\footnote{\url{https://swayam.gov.in/}} and NPTEL\footnote{\url{https://nptel.ac.in/}} platforms. For medical data, the corpus included content from various medical research protocols for research studies, such as consent forms, information sheets, and medical awareness materials. These were obtained from a well-known private Christian minority community-run medical college and hospital. We first transcribed these video contents verbatim with the help of respective language experts, ensuring accurate transcription. Subsequently, post-editing was performed on machine-translated content generated using multiple translation engines, such as SSMT\footnote{\url{http://ssmt.iiit.ac.in/translate}} and Google Translate\footnote{\url{https://translate.google.co.in/}}. Given the technical nature of the task, we provided clear guidelines to the translators to ensure consistent quality and adherence to domain standards. The guidelines included:

\begin{itemize}
    \justifying
    \item \textbf{Thorough Understanding:} Translators need to carefully read the entire source text before beginning the translation process to grasp its meaning and context fully.  
    \item \textbf{Faithfulness to the Source} Translations need to accurately capture the meaning of each sentence and remain as clear and understandable as the original text as a whole.  
    \item \textbf{Rigorous Review} It is advised to review own work multiple times, both silently and aloud, to ensure clarity, coherence, and appropriateness of word choice.  
    \item \textbf{Consistency in Terminology} Technical terms and expressions are have to be used consistently throughout the translation. Translators need to rely on established term translations present in resources like NCERT textbooks\footnote{\url{https://ncert.nic.in/textbook.php}}. If a term lacked a clear translation, it is to be transliterated.  
    \item \textbf{Contextual Alignment} The overall translation has to convey the same natural message and context as the original text.  
\end{itemize}

To ensure quality, an in-house team of experienced translators (4 to 5 experts for each language pair) rigorously validated each sentence translated by freelancers following the same guidelines. Discrepancies or deviations from the guidelines are corrected by reviewers as part of the process. All these activities are conducted using a custom-built collaborative language platform, PostEditMe\footnote{\url{https://posteditme.in/}}, which effectively managed workflows for translation, post-editing, validation, and task assignments throughout this effort. The developed human post-edited parallel corpora contain healthcare and educational domains spanning over a wide range of sub-domains, including Psychology, Natural Sciences, Teaching Methods, Mathematics, Political Science, History, Communication Skills, Law, Tourism, Computer Science, Marketing, Management, Textile, Economics, and Health, ensuring comprehensive coverage of diverse academic disciplines. Notably, the post-edited parallel corpora were developed with English as the source language, and the same English text was post-edited into other Indian languages, enabling the creation of n-way parallel corpora among Indian languages.

\subsection*{Quality Synthetic Parallel Corpora}

Synthetic corpora can address the paucity of parallel data for Indian languages. To bridge this gap, we utilized various techniques, including pivot-based translation, back-translation, and forward translation at the sentence and paragraph level. These methods significantly enhance the scope of existing corpora, expanding their coverage and offering vital support for low-resource language pairs.

\subsubsection*{Pivoted Parallel Corpora} Most of the parallel corpora developed for Indian languages focus on translations involving English, with relatively few efforts dedicated to creating corpora for Indian-to-Indian language translations. However, such corpora are crucial for addressing the linguistic diversity and translation needs of the Indian subcontinent. Pivot-based translation offers an effective solution to the lack of direct parallel corpora for underrepresented language pairs. \\

In this work, we enrich Indian-to-Indian language corpora by generating high-quality synthetic data using pivot translation. Our approach uses well-represented languages such as English and Hindi as intermediary "pivot" languages to bridge the gap between other low-resource Indian languages. We leveraged existing corpora for these languages and employed SSMT Translator\footnote{\url{http://ssmt.iiit.ac.in/translatev3}} to translate from Indian languages to English, and then from English to other Indian languages under consideration. Similarly, translations are generated between Indian languages and Hindi, using Hindi as a pivot. This process involves creating automatic translations first between Indian languages and the pivot (English or Hindi), followed by translations from the pivot to the target Indian languages. \\

To address the varying quality of generated parallel corpora, we implemented filtering mechanisms to ensure the reliability of the training data. Advanced methods, such as COMET-QE \citep{rei2022comet}, were employed to evaluate the quality of sentence pairs by assigning quality estimation scores. Sentence pairs falling below a set threshold (determined as the average QE score across all pairs for each language pair) were filtered out, maintaining a high standard for the training datasets. To further validate this approach, randomly selected sentences from the filtered corpora were reviewed by in-house language experts. Their evaluations confirmed that the majority of the automatically scored and filtered data were valid parallel corpora, demonstrating its effectiveness in supporting machine translation models and enhancing their performance and generalizability. By leveraging existing translations involving the pivot languages, this method enables the generation of new parallel corpora, thereby extending the reach of machine translation models to low-resource language pairs. This approach facilitates cross-lingual applications and fosters inclusivity in multilingual natural language processing.

\subsubsection*{Iterative Backward Translation} We employed back-translation, a widely used data augmentation technique in machine translation, to enhance the diversity and quality of training data. Monolingual corpora for the target languages were sourced from WMT News Crawl\footnote{\url{https://data.statmt.org/news-crawl/}} and Wikipedia\footnote{\url{https://huggingface.co/datasets/wikimedia/wikipedia}}. Using SSMT Translator\footnote{\url{http://ssmt.iiit.ac.in/translatev3}} and IndicTrans2\footnote{\url{https://github.com/AI4Bharat/IndicTrans2}}, we translated this monolingual data bidirectionally: from Indian languages to English and from English to Indian languages. The translated data introduced variations in sentence structure, vocabulary, and phrasing, enriching the training set's diversity. For each monolingual corpus, both translation outputs were evaluated using COMET-QE scores, and the version with the highest score is retained. This filtered data is then used to train a multilingual machine translation model involving the selected language pairs. The newly trained model is subsequently used to re-translate the same monolingual corpora, iteratively improving the translations. This process is repeated for five iterations, with the best-scored translation pairs from all iterations being selected for the final training corpus.

\paragraph{\textbf{Paragraph Level Backward Translation}} For the Wikipedia dataset\footnote{\url{https://huggingface.co/datasets/wikimedia/wikipedia}} and crawled news articles from the internet and from the available monolingual corpora\footnote{\url{https://data.statmt.org/news-crawl/}}, we adopted paragraph-level backward translation to preserve coherence and contextual consistency across entire paragraphs. Unlike sentence-level back-translation, this approach ensures that the translated content retains discourse-level attributes, such as topic continuity, pronoun resolution, and lexical cohesion. By focusing on paragraph-level fidelity, this method enhances the quality of machine translation systems, particularly for tasks requiring deeper contextual understanding, such as document translation or conversational AI. The same filtering mechanism is applied here as well. Each sentence within a paragraph is evaluated for quality using COMET-QE scores, and the best-scored translations were selected. These high-quality sentence translations were then reassembled into paragraphs to maintain the original discourse structure. This strategy ensured the development of a robust multilingual machine translation model capable of handling complex and contextually rich translations.

\subsection*{Parallel Corpora Cleaning}

Cleaning parallel corpora is crucial for ensuring the accuracy and reliability of training datasets. Noise in parallel corpora, such as duplicates, incomplete translations, misalignments, script variations, and unwanted symbols, can significantly degrade the performance of machine translation (MT) models. To address these issues, we utilized automated tools to detect and resolve errors or remove problematic sentence pairs. This process involved normalizing text and tackling script-specific challenges in Indian languages, such as script diversity and spelling inconsistencies. Additional preprocessing steps included removing mistranslations, addressing code-mixing, and correcting tokenization errors. We employed multiple methodologies to achieve effective corpus cleaning:

\paragraph{Length-Based Filtering}  
This method filters parallel corpora by comparing differences in word and character counts to ensure alignment quality. For example, we calculate the average sentence length for each language and maintain an acceptable range of differences for a given language pair in terms of words and characters. This range, set as the threshold, allows for a variation of ±10 words (or corresponding characters). Every source and target text pair in the generated corpora is evaluated against this threshold. Sentence pairs that do not meet this criterion are discarded to maintain the quality of the dataset.

\paragraph{Language and Script Identification}  
We utilized a FastText-based language detection tool to identify the language of each word within a sentence. If the majority of word-level language tags do not align with the sentence-level language tag, the sentence pair is filtered out to ensure linguistic consistency.  

\paragraph{HTML/XML Tag Validation}  
Sentence pairs with mismatched HTML or XML tags between the source and target sides are removed to avoid structural inconsistencies in the data. 

\paragraph{COMET-QE Scoring}  
The quality of source-target and target-source sentence pairs is evaluated using COMET-22 scores. Sentence pairs with scores significantly below the average threshold (average minus 10) are excluded, ensuring that only high-quality data is retained for training.  \\

We applied these cleaning methods to existing, newly developed, and synthetically generated parallel corpora for Indian languages to achieve greater alignment accuracy, noise reduction, and linguistic consistency, resulting in more effective MT systems.

\subsection{Developed Parallel Corpora}

\begin{table}[H]
\caption{The generated over 1B parallel corpora cover 325 language pairs, encompassing 36 languages from the Indian subcontinent and utilizing 15 unique scripts.}
\label{tab:bhashik-parallel-corpora-generic}
\resizebox{\columnwidth}{!}{%
\begin{tabular}{lr|lr|lr|lr|lr|lr|ll|}
\textbf{Language Pair} & \multicolumn{1}{l|}{\textbf{\begin{tabular}[c]{@{}l@{}}Developed \\ Parallel Data\end{tabular}}} & \textbf{Language Pair} & \multicolumn{1}{l|}{\textbf{\begin{tabular}[c]{@{}l@{}}Developed \\ Parallel Data\end{tabular}}} & \textbf{Language Pair} & \multicolumn{1}{l|}{\textbf{\begin{tabular}[c]{@{}l@{}}Developed \\ Parallel Data\end{tabular}}} & \textbf{Language Pair} & \multicolumn{1}{l|}{\textbf{\begin{tabular}[c]{@{}l@{}}Developed \\ Parallel Data\end{tabular}}} & \textbf{Language Pair} & \multicolumn{1}{l|}{\textbf{\begin{tabular}[c]{@{}l@{}}Developed \\ Parallel Data\end{tabular}}} & \textbf{Language Pair} & \multicolumn{1}{l|}{\textbf{\begin{tabular}[c]{@{}l@{}}Developed \\ Parallel Data\end{tabular}}} & \textbf{Language Pair} & \textbf{\begin{tabular}[c]{@{}l@{}}Developed \\ Parallel Data\end{tabular}} \\ \hline
asm\_Beng\_ben\_Beng   & 2111717                                                                                          & brx\_Deva\_eng\_Latn   & 21161986                                                                                         & eng\_Latn\_mai\_Deva   & 25730422                                                                                         & guj\_Gujr\_snd\_Deva   & 7948035                                                                                          & kas\_Arab\_snd\_Arab   & 872206                                                                                           & mar\_Deva\_ory\_Orya   & 47036189                                                                                         & pan\_Guru\_snd\_Deva   & \multicolumn{1}{r|}{3246797}                                                \\
asm\_Beng\_brx\_Deva   & 765615                                                                                           & brx\_Deva\_gom\_Deva   & 799906                                                                                           & eng\_Latn\_mal\_Mlym   & 103881922                                                                                        & guj\_Gujr\_tam\_Taml   & 69922990                                                                                         & kas\_Arab\_snd\_Deva   & 1231344                                                                                          & mar\_Deva\_pan\_Guru   & 4421324                                                                                          & pan\_Guru\_tam\_Taml   & \multicolumn{1}{r|}{4207666}                                                \\
asm\_Beng\_doi\_Deva   & 2317597                                                                                          & brx\_Deva\_guj\_Gujr   & 3423331                                                                                          & eng\_Latn\_mar\_Deva   & 98133618                                                                                         & guj\_Gujr\_tel\_Telu   & 635961470                                                                                        & kas\_Arab\_tam\_Taml   & 2359634                                                                                          & mar\_Deva\_san\_Deva   & 1940249                                                                                          & pan\_Guru\_tel\_Telu   & \multicolumn{1}{r|}{80308819}                                               \\
asm\_Beng\_eng\_Latn   & 30594977                                                                                         & brx\_Deva\_hin\_Deva   & 7883998                                                                                          & eng\_Latn\_mni\_Beng   & 13473789                                                                                         & guj\_Gujr\_urd\_Arab   & 171350512                                                                                        & kas\_Arab\_tel\_Telu   & 10578313                                                                                         & mar\_Deva\_sat\_Olck   & 539452                                                                                           & pan\_Guru\_urd\_Arab   & \multicolumn{1}{r|}{7873795}                                                \\
asm\_Beng\_gom\_Deva   & 828224                                                                                           & brx\_Deva\_kan\_Knda   & 2247979                                                                                          & eng\_Latn\_mni\_Mtei   & 24917741                                                                                         & hin\_Deva\_kan\_Knda   & 27641372                                                                                         & kas\_Arab\_urd\_Arab   & 3362396                                                                                          & mar\_Deva\_snd\_Arab   & 2047746                                                                                          & san\_Deva\_sat\_Olck   & \multicolumn{1}{r|}{355047}                                                 \\
asm\_Beng\_guj\_Gujr   & 9163504                                                                                          & brx\_Deva\_kas\_Arab   & 710060                                                                                           & eng\_Latn\_npi\_Deva   & 79094284                                                                                         & hin\_Deva\_kas\_Arab   & 10689877                                                                                         & kas\_Deva\_mai\_Deva   & 1723270                                                                                          & mar\_Deva\_snd\_Deva   & 2682147                                                                                          & san\_Deva\_snd\_Arab   & \multicolumn{1}{r|}{1135478}                                                \\
asm\_Beng\_hin\_Deva   & 8361026                                                                                          & brx\_Deva\_kas\_Deva   & 1429735                                                                                          & eng\_Latn\_ory\_Orya   & 434864407                                                                                        & hin\_Deva\_kas\_Deva   & 18466409                                                                                         & kas\_Deva\_mal\_Mlym   & 3812953                                                                                          & mar\_Deva\_tam\_Taml   & 3041800                                                                                          & san\_Deva\_snd\_Deva   & \multicolumn{1}{r|}{1453221}                                                \\
asm\_Beng\_kan\_Knda   & 2374887                                                                                          & brx\_Deva\_mai\_Deva   & 780823                                                                                           & eng\_Latn\_pan\_Guru   & 106211275                                                                                        & hin\_Deva\_mai\_Deva   & 8774341                                                                                          & kas\_Deva\_mar\_Deva   & 3342699                                                                                          & mar\_Deva\_tel\_Telu   & 79927241                                                                                         & san\_Deva\_tam\_Taml   & \multicolumn{1}{r|}{2085085}                                                \\
asm\_Beng\_kas\_Arab   & 880736                                                                                           & brx\_Deva\_mal\_Mlym   & 1986655                                                                                          & eng\_Latn\_san\_Deva   & 43354388                                                                                         & hin\_Deva\_mal\_Mlym   & 81376763                                                                                         & kas\_Deva\_mni\_Beng   & 819979                                                                                           & mar\_Deva\_urd\_Arab   & 40450921                                                                                         & san\_Deva\_tel\_Telu   & \multicolumn{1}{r|}{29210615}                                               \\
asm\_Beng\_kas\_Deva   & 1577560                                                                                          & brx\_Deva\_mar\_Deva   & 1462688                                                                                          & eng\_Latn\_sat\_Olck   & 9625012                                                                                          & hin\_Deva\_mar\_Deva   & 80444061                                                                                         & kas\_Deva\_mni\_Mtei   & 1503829                                                                                          & mni\_Beng\_mni\_Mtei   & 356775                                                                                           & san\_Deva\_urd\_Arab   & \multicolumn{1}{r|}{3636643}                                                \\
asm\_Beng\_mai\_Deva   & 757165                                                                                           & brx\_Deva\_mni\_Beng   & 378232                                                                                           & eng\_Latn\_snd\_Arab   & 31294443                                                                                         & hin\_Deva\_mni\_Beng   & 3728206                                                                                          & kas\_Deva\_npi\_Deva   & 2457536                                                                                          & mni\_Beng\_npi\_Deva   & 389988                                                                                           & sat\_Olck\_snd\_Arab   & \multicolumn{1}{r|}{348940}                                                 \\
asm\_Beng\_mal\_Mlym   & 1776146                                                                                          & brx\_Deva\_mni\_Mtei   & 758003                                                                                           & eng\_Latn\_snd\_Deva   & 31606531                                                                                         & hin\_Deva\_mni\_Mtei   & 9416582                                                                                          & kas\_Deva\_ory\_Orya   & 19983032                                                                                         & mni\_Beng\_ory\_Orya   & 3585322                                                                                          & sat\_Olck\_snd\_Deva   & \multicolumn{1}{r|}{489121}                                                 \\
asm\_Beng\_mar\_Deva   & 1418576                                                                                          & brx\_Deva\_npi\_Deva   & 954826                                                                                           & eng\_Latn\_tam\_Taml   & 94940774                                                                                         & hin\_Deva\_npi\_Deva   & 10786012                                                                                         & kas\_Deva\_pan\_Guru   & 4201119                                                                                          & mni\_Beng\_pan\_Guru   & 1119674                                                                                          & sat\_Olck\_tam\_Taml   & \multicolumn{1}{r|}{430418}                                                 \\
asm\_Beng\_mni\_Beng   & 312229                                                                                           & brx\_Deva\_ory\_Orya   & 7615181                                                                                          & eng\_Latn\_tel\_Telu   & 960415291                                                                                        & hin\_Deva\_ory\_Orya   & 346421975                                                                                        & kas\_Deva\_san\_Deva   & 1941751                                                                                          & mni\_Beng\_san\_Deva   & 525579                                                                                           & sat\_Olck\_tel\_Telu   & \multicolumn{1}{r|}{2579655}                                                \\
asm\_Beng\_mni\_Mtei   & 888899                                                                                           & brx\_Deva\_pan\_Guru   & 2189039                                                                                          & eng\_Latn\_urd\_Arab   & 246732366                                                                                        & hin\_Deva\_pan\_Guru   & 63333450                                                                                         & kas\_Deva\_sat\_Olck   & 676632                                                                                           & mni\_Beng\_sat\_Olck   & 139996                                                                                           & sat\_Olck\_urd\_Arab   & \multicolumn{1}{r|}{976839}                                                 \\
asm\_Beng\_npi\_Deva   & 1009181                                                                                          & brx\_Deva\_san\_Deva   & 751939                                                                                           & gom\_Deva\_guj\_Gujr   & 9962240                                                                                          & hin\_Deva\_san\_Deva   & 29108161                                                                                         & kas\_Deva\_snd\_Arab   & 1837653                                                                                          & mni\_Beng\_snd\_Arab   & 609824                                                                                           & snd\_Arab\_snd\_Deva   & \multicolumn{1}{r|}{1292311}                                                \\
asm\_Beng\_ory\_Orya   & 12801566                                                                                         & brx\_Deva\_sat\_Olck   & 240046                                                                                           & gom\_Deva\_hin\_Deva   & 13998575                                                                                         & hin\_Deva\_sat\_Olck   & 3377744                                                                                          & kas\_Deva\_snd\_Deva   & 1827310                                                                                          & mni\_Beng\_snd\_Deva   & 659939                                                                                           & snd\_Arab\_tam\_Taml   & \multicolumn{1}{r|}{1890070}                                                \\
asm\_Beng\_pan\_Guru   & 2110174                                                                                          & brx\_Deva\_snd\_Arab   & 900441                                                                                           & gom\_Deva\_kan\_Knda   & 2432914                                                                                          & hin\_Deva\_snd\_Arab   & 18584708                                                                                         & kas\_Deva\_tam\_Taml   & 3329492                                                                                          & mni\_Beng\_tam\_Taml   & 719910                                                                                           & snd\_Arab\_tel\_Telu   & \multicolumn{1}{r|}{19075895}                                               \\
asm\_Beng\_san\_Deva   & 1034015                                                                                          & brx\_Deva\_snd\_Deva   & 1060504                                                                                          & gom\_Deva\_kas\_Arab   & 907722                                                                                           & hin\_Deva\_snd\_Deva   & 14552624                                                                                         & kas\_Deva\_tel\_Telu   & 19804719                                                                                         & mni\_Beng\_tel\_Telu   & 2192230                                                                                          & snd\_Arab\_urd\_Arab   & \multicolumn{1}{r|}{3048720}                                                \\
asm\_Beng\_sat\_Olck   & 439289                                                                                           & brx\_Deva\_tam\_Taml   & 1321496                                                                                          & gom\_Deva\_kas\_Deva   & 1739739                                                                                          & hin\_Deva\_tam\_Taml   & 76095904                                                                                         & kas\_Deva\_urd\_Arab   & 5387510                                                                                          & mni\_Beng\_urd\_Arab   & 1254597                                                                                          & snd\_Deva\_tam\_Taml   & \multicolumn{1}{r|}{2661636}                                                \\
asm\_Beng\_snd\_Arab   & 929278                                                                                           & brx\_Deva\_tel\_Telu   & 6450144                                                                                          & gom\_Deva\_mai\_Deva   & 760450                                                                                           & hin\_Deva\_tel\_Telu   & 854298767                                                                                        & mai\_Deva\_mal\_Mlym   & 2096537                                                                                          & mni\_Mtei\_npi\_Deva   & 1443672                                                                                          & snd\_Deva\_tel\_Telu   & \multicolumn{1}{r|}{15779583}                                               \\
asm\_Beng\_snd\_Deva   & 1315760                                                                                          & brx\_Deva\_urd\_Arab   & 2250715                                                                                          & gom\_Deva\_mal\_Mlym   & 2020062                                                                                          & hin\_Deva\_urd\_Arab   & 211718893                                                                                        & mai\_Deva\_mar\_Deva   & 1616046                                                                                          & mni\_Mtei\_ory\_Orya   & 9403109                                                                                          & snd\_Deva\_urd\_Arab   & \multicolumn{1}{r|}{4222851}                                                \\
asm\_Beng\_tam\_Taml   & 1645025                                                                                          & doi\_Deva\_eng\_Latn   & 56050632                                                                                         & gom\_Deva\_mar\_Deva   & 1393256                                                                                          & kan\_Knda\_kas\_Arab   & 2720541                                                                                          & mai\_Deva\_mni\_Beng   & 549939                                                                                           & mni\_Mtei\_pan\_Guru   & 2435659                                                                                          & tam\_Taml\_tel\_Telu   & \multicolumn{1}{r|}{74713991}                                               \\
asm\_Beng\_tel\_Telu   & 17575936                                                                                         & doi\_Deva\_gom\_Deva   & 2230764                                                                                          & gom\_Deva\_mni\_Beng   & 389990                                                                                           & kan\_Knda\_kas\_Deva   & 4425169                                                                                          & mai\_Deva\_mni\_Mtei   & 849017                                                                                           & mni\_Mtei\_san\_Deva   & 1042586                                                                                          & tam\_Taml\_urd\_Arab   & \multicolumn{1}{r|}{22118281}                                               \\
asm\_Beng\_urd\_Arab   & 2786265                                                                                          & doi\_Deva\_guj\_Gujr   & 13475348                                                                                         & gom\_Deva\_mni\_Mtei   & 1025298                                                                                          & kan\_Knda\_mai\_Deva   & 2465368                                                                                          & mai\_Deva\_npi\_Deva   & 1095355                                                                                          & mni\_Mtei\_sat\_Olck   & 329603                                                                                           & tel\_Telu\_urd\_Arab   & \multicolumn{1}{r|}{242494781}                                              \\
ben\_Beng\_brx\_Deva   & 1771330                                                                                          & doi\_Deva\_hin\_Deva   & 25314841                                                                                         & gom\_Deva\_npi\_Deva   & 1344302                                                                                          & kan\_Knda\_mal\_Mlym   & 6077765                                                                                          & mai\_Deva\_ory\_Orya   & 8857273                                                                                          & mni\_Mtei\_snd\_Arab   & 1156456                                                                                          &                        &                                                                             \\
ben\_Beng\_doi\_Deva   & 5725017                                                                                          & doi\_Deva\_kan\_Knda   & 6225786                                                                                          & gom\_Deva\_ory\_Orya   & 11288994                                                                                         & kan\_Knda\_mar\_Deva   & 4775780                                                                                          & mai\_Deva\_pan\_Guru   & 2318762                                                                                          & mni\_Mtei\_snd\_Deva   & 1302097                                                                                          &                        &                                                                             \\
ben\_Beng\_eng\_Latn   & 110423482                                                                                        & doi\_Deva\_kas\_Arab   & 2260841                                                                                          & gom\_Deva\_pan\_Guru   & 1984080                                                                                          & kan\_Knda\_mni\_Beng   & 1409277                                                                                          & mai\_Deva\_san\_Deva   & 1043353                                                                                          & mni\_Mtei\_tam\_Taml   & 1748302                                                                                          &                        &                                                                             \\
ben\_Beng\_gom\_Deva   & 2152903                                                                                          & doi\_Deva\_kas\_Deva   & 3668955                                                                                          & gom\_Deva\_san\_Deva   & 942936                                                                                           & kan\_Knda\_mni\_Mtei   & 2604263                                                                                          & mai\_Deva\_sat\_Olck   & 360250                                                                                           & mni\_Mtei\_tel\_Telu   & 8112526                                                                                          &                        &                                                                             \\
ben\_Beng\_guj\_Gujr   & 78967455                                                                                         & doi\_Deva\_mai\_Deva   & 2190770                                                                                          & gom\_Deva\_sat\_Olck   & 399791                                                                                           & kan\_Knda\_npi\_Deva   & 3973567                                                                                          & mai\_Deva\_snd\_Arab   & 1022691                                                                                          & mni\_Mtei\_urd\_Arab   & 2734445                                                                                          &                        &                                                                             \\
ben\_Beng\_hin\_Deva   & 87189520                                                                                         & doi\_Deva\_mal\_Mlym   & 5447712                                                                                          & gom\_Deva\_snd\_Arab   & 931740                                                                                           & kan\_Knda\_ory\_Orya   & 55381962                                                                                         & mai\_Deva\_snd\_Deva   & 1121301                                                                                          & npi\_Deva\_ory\_Orya   & 11198582                                                                                         &                        &                                                                             \\
ben\_Beng\_kan\_Knda   & 6588212                                                                                          & doi\_Deva\_mar\_Deva   & 4181626                                                                                          & gom\_Deva\_snd\_Deva   & 1120965                                                                                          & kan\_Knda\_pan\_Guru   & 6601794                                                                                          & mai\_Deva\_tam\_Taml   & 1780322                                                                                          & npi\_Deva\_pan\_Guru   & 3319821                                                                                          &                        &                                                                             \\
ben\_Beng\_kas\_Arab   & 2145259                                                                                          & doi\_Deva\_mni\_Beng   & 1279927                                                                                          & gom\_Deva\_tam\_Taml   & 1718214                                                                                          & kan\_Knda\_san\_Deva   & 3279907                                                                                          & mai\_Deva\_tel\_Telu   & 11267803                                                                                         & npi\_Deva\_san\_Deva   & 1565423                                                                                          &                        &                                                                             \\
ben\_Beng\_kas\_Deva   & 4232603                                                                                          & doi\_Deva\_mni\_Mtei   & 2217101                                                                                          & gom\_Deva\_tel\_Telu   & 13826104                                                                                         & kan\_Knda\_sat\_Olck   & 928128                                                                                           & mai\_Deva\_urd\_Arab   & 2769033                                                                                          & npi\_Deva\_sat\_Olck   & 360263                                                                                           &                        &                                                                             \\
ben\_Beng\_mai\_Deva   & 2254028                                                                                          & doi\_Deva\_npi\_Deva   & 3384768                                                                                          & gom\_Deva\_urd\_Arab   & 5767926                                                                                          & kan\_Knda\_snd\_Arab   & 2572544                                                                                          & mal\_Mlym\_mar\_Deva   & 4080152                                                                                          & npi\_Deva\_snd\_Arab   & 1382765                                                                                          &                        &                                                                             \\
ben\_Beng\_mal\_Mlym   & 5151355                                                                                          & doi\_Deva\_ory\_Orya   & 26264311                                                                                         & guj\_Gujr\_hin\_Deva   & 501919187                                                                                        & kan\_Knda\_snd\_Deva   & 3701804                                                                                          & mal\_Mlym\_mni\_Beng   & 1005011                                                                                          & npi\_Deva\_snd\_Deva   & 2383049                                                                                          &                        &                                                                             \\
ben\_Beng\_mar\_Deva   & 4356579                                                                                          & doi\_Deva\_pan\_Guru   & 6035401                                                                                          & guj\_Gujr\_kan\_Knda   & 45730533                                                                                         & kan\_Knda\_tam\_Taml   & 4934229                                                                                          & mal\_Mlym\_mni\_Mtei   & 1994893                                                                                          & npi\_Deva\_tam\_Taml   & 1780516                                                                                          &                        &                                                                             \\
ben\_Beng\_mni\_Beng   & 897884                                                                                           & doi\_Deva\_san\_Deva   & 2737805                                                                                          & guj\_Gujr\_kas\_Arab   & 5266708                                                                                          & kan\_Knda\_tel\_Telu   & 90235065                                                                                         & mal\_Mlym\_npi\_Deva   & 2336955                                                                                          & npi\_Deva\_tel\_Telu   & 64765229                                                                                         &                        &                                                                             \\
ben\_Beng\_mni\_Mtei   & 2310198                                                                                          & doi\_Deva\_sat\_Olck   & 849840                                                                                           & guj\_Gujr\_kas\_Deva   & 10007355                                                                                         & kan\_Knda\_urd\_Arab   & 8070849                                                                                          & mal\_Mlym\_ory\_Orya   & 50442690                                                                                         & npi\_Deva\_urd\_Arab   & 3595423                                                                                          &                        &                                                                             \\
ben\_Beng\_npi\_Deva   & 2621108                                                                                          & doi\_Deva\_snd\_Arab   & 2211042                                                                                          & guj\_Gujr\_mai\_Deva   & 4852703                                                                                          & kas\_Arab\_kas\_Deva   & 1799571                                                                                          & mal\_Mlym\_pan\_Guru   & 5425283                                                                                          & ory\_Orya\_pan\_Guru   & 48738324                                                                                         &                        &                                                                             \\
ben\_Beng\_ory\_Orya   & 52347364                                                                                         & doi\_Deva\_snd\_Deva   & 2821105                                                                                          & guj\_Gujr\_mal\_Mlym   & 41472075                                                                                         & kas\_Arab\_mai\_Deva   & 1140655                                                                                          & mal\_Mlym\_san\_Deva   & 2397623                                                                                          & ory\_Orya\_san\_Deva   & 20414334                                                                                         &                        &                                                                             \\
ben\_Beng\_pan\_Guru   & 5688148                                                                                          & doi\_Deva\_tam\_Taml   & 4419242                                                                                          & guj\_Gujr\_mar\_Deva   & 76034043                                                                                         & kas\_Arab\_mal\_Mlym   & 2320653                                                                                          & mal\_Mlym\_sat\_Olck   & 713045                                                                                           & ory\_Orya\_sat\_Olck   & 3094866                                                                                          &                        &                                                                             \\
ben\_Beng\_san\_Deva   & 2876914                                                                                          & doi\_Deva\_tel\_Telu   & 26541111                                                                                         & guj\_Gujr\_mni\_Beng   & 1289734                                                                                          & kas\_Arab\_mar\_Deva   & 2033332                                                                                          & mal\_Mlym\_snd\_Arab   & 2369219                                                                                          & ory\_Orya\_snd\_Arab   & 14829766                                                                                         &                        &                                                                             \\
ben\_Beng\_sat\_Olck   & 899756                                                                                           & doi\_Deva\_urd\_Arab   & 6725444                                                                                          & guj\_Gujr\_mni\_Mtei   & 4373466                                                                                          & kas\_Arab\_mni\_Beng   & 439988                                                                                           & mal\_Mlym\_snd\_Deva   & 3280861                                                                                          & ory\_Orya\_snd\_Deva   & 15923972                                                                                         &                        &                                                                             \\
ben\_Beng\_snd\_Arab   & 2322747                                                                                          & eng\_Latn\_gom\_Deva   & 27718817                                                                                         & guj\_Gujr\_npi\_Deva   & 30603002                                                                                         & kas\_Arab\_mni\_Mtei   & 928839                                                                                           & mal\_Mlym\_tam\_Taml   & 4176661                                                                                          & ory\_Orya\_tam\_Taml   & 43120335                                                                                         &                        &                                                                             \\
ben\_Beng\_snd\_Deva   & 3155804                                                                                          & eng\_Latn\_guj\_Gujr   & 596740761                                                                                        & guj\_Gujr\_ory\_Orya   & 217767278                                                                                        & kas\_Arab\_npi\_Deva   & 1384843                                                                                          & mal\_Mlym\_tel\_Telu   & 81261194                                                                                         & ory\_Orya\_tel\_Telu   & 430745670                                                                                        &                        &                                                                             \\
ben\_Beng\_tam\_Taml   & 4180758                                                                                          & eng\_Latn\_hin\_Deva   & 832738952                                                                                        & guj\_Gujr\_pan\_Guru   & 40878961                                                                                         & kas\_Arab\_ory\_Orya   & 10958432                                                                                         & mal\_Mlym\_urd\_Arab   & 6862158                                                                                          & ory\_Orya\_urd\_Arab   & 61149573                                                                                         &                        &                                                                             \\
ben\_Beng\_tel\_Telu   & 87561402                                                                                         & eng\_Latn\_kan\_Knda   & 115521466                                                                                        & guj\_Gujr\_san\_Deva   & 14827663                                                                                         & kas\_Arab\_pan\_Guru   & 2423309                                                                                          & mar\_Deva\_mni\_Beng   & 869866                                                                                           & pan\_Guru\_san\_Deva   & 2766113                                                                                          &                        &                                                                             \\
ben\_Beng\_urd\_Arab   & 42602867                                                                                         & eng\_Latn\_kas\_Arab   & 25142330                                                                                         & guj\_Gujr\_sat\_Olck   & 1541029                                                                                          & kas\_Arab\_san\_Deva   & 1123349                                                                                          & mar\_Deva\_mni\_Mtei   & 1763291                                                                                          & pan\_Guru\_sat\_Olck   & 840630                                                                                           &                        &                                                                             \\
brx\_Deva\_doi\_Deva   & 1916922                                                                                          & eng\_Latn\_kas\_Deva   & 41442708                                                                                         & guj\_Gujr\_snd\_Arab   & 9319233                                                                                          & kas\_Arab\_sat\_Olck   & 459615                                                                                           & mar\_Deva\_npi\_Deva   & 1783813                                                                                          & pan\_Guru\_snd\_Arab   & 2625050                                                                                          &                        &                                                                             \\
\end{tabular}%
}
\end{table}
\begin{table}[]
\caption{Postedited Health Domain Corpora for English and 8 Indian Languages}
\label{tab:bhashik-parallel-corpora-domain-health}
\resizebox{\textwidth}{!}{%
\begin{tabular}{ll|ll}
\textbf{Language Pair} & \textbf{Passages/Sentences} & \textbf{Language Pair} & \textbf{Passages/Sentences} \\ \hline
ben\_Beng\_Latn-tam\_Taml & 142 & eng\_Latn-tel\_Telu & 10308 \\
ben\_Beng\_Latn-tel\_Telu & 291 & hin\_Deva-ben\_Beng\_Latn & 6384 \\
eng\_Latn-ben\_Beng\_Latn & 7915 & hin\_Deva-guj\_Gujr & 145 \\
eng\_Latn-guj\_Gujr & 158 & hin\_Deva-kan\_Knda & 1138 \\
eng\_Latn-hin\_Deva & 11042 & hin\_Deva-mal\_Mlym & 1436 \\
eng\_Latn-kan\_Knda & 1209 & hin\_Deva-mar\_Deva & 879 \\
eng\_Latn-mal\_Mlym & 1754 & hin\_Deva-tam\_Taml & 5517 \\
eng\_Latn-mar\_Deva & 906 & hin\_Deva-tel\_Telu & 6412 \\
eng\_Latn-tam\_Taml & 8726 & tam\_Taml-tel\_Telu & 1019
\end{tabular}%
}
\end{table}
\begin{table}[]
\caption{The human post-edited 2M domain-parallel corpora spans English and 10 Indian languages, covering 15 educational domains.}
\label{tab:bhashik-parallel-corpora-domain}
\resizebox{\columnwidth}{!}{%
\begin{tabular}{llr|llr}
\hline
\textbf{Domain}  & \textbf{Language Pair} & \multicolumn{1}{l|}{\textbf{Passages/Sentences}} & \textbf{Domain}     & \textbf{Language Pair} & \multicolumn{1}{l}{\textbf{Passages/Sentences}} \\ \hline
Natural science  & eng\_Latn-hin\_Deva    & 140999                              & Teaching methods    & eng\_Latn-tam\_Taml    & 17                                 \\
Natural science  & eng\_Latn-mar\_Deva    & 191936                              & Teaching methods    & eng\_Latn-mal\_Mlym    & 527                                \\
Natural science  & eng\_Latn-kan\_Knda    & 185491                              & Teaching methods    & eng\_Latn-urd\_Arab    & 3909                               \\
Natural science  & eng\_Latn-tel\_Telu    & 223146                              & Management          & eng\_Latn-mar\_Deva    & 14521                              \\
Natural science  & eng\_Latn-mal\_Mlym    & 84                                  & Management          & eng\_Latn-kan\_Knda    & 11001                              \\
Natural science  & eng\_Latn-tam\_Taml    & 38                                  & Management          & eng\_Latn-tel\_Telu    & 11041                              \\
Computer science & eng\_Latn-mar\_Deva    & 80170                               & Health              & eng\_Latn-hin\_Deva    & 715                                \\
Computer science & eng\_Latn-hin\_Deva    & 71507                               & Psychology          & eng\_Latn-hin\_Deva    & 12651                              \\
Computer science & eng\_Latn-kan\_Knda    & 75354                               & Psychology          & eng\_Latn-tel\_Telu    & 12771                              \\
Computer science & eng\_Latn-tel\_Telu    & 75388                               & Psychology          & eng\_Latn-mar\_Deva    & 12822                              \\
Computer science & eng\_Latn-urd\_Arab    & 332                                 & Psychology          & eng\_Latn-kan\_Knda    & 12570                              \\
Computer science & eng\_Latn-tam\_Taml    & 11902                               & Marketing           & eng\_Latn-hin\_Deva    & 4903                               \\
Law              & eng\_Latn-tel\_Telu    & 80165                               & Marketing           & eng\_Latn-tel\_Telu    & 4971                               \\
Law              & eng\_Latn-mar\_Deva    & 81581                               & Marketing           & eng\_Latn-mar\_Deva    & 4858                               \\
Law              & eng\_Latn-tam\_Taml    & 90                                  & Marketing           & eng\_Latn-kan\_Knda    & 5099                               \\
Law              & eng\_Latn-kan\_Knda    & 78245                               & Political science   & eng\_Latn-tel\_Telu    & 11978                              \\
Law              & eng\_Latn-hin\_Deva    & 735                                 & Political science   & eng\_Latn-mar\_Deva    & 11935                              \\
Mathematics      & eng\_Latn-mar\_Deva    & 62881                               & Political science   & eng\_Latn-kan\_Knda    & 12001                              \\
Mathematics      & eng\_Latn-tel\_Telu    & 59373                               & Political science   & eng\_Latn-hin\_Deva    & 7876                               \\
Mathematics      & eng\_Latn-kan\_Knda    & 40625                               & Textile             & eng\_Latn-tel\_Telu    & 7609                               \\
Mathematics      & eng\_Latn-hin\_Deva    & 26259                               & Textile             & eng\_Latn-mar\_Deva    & 7637                               \\
Mathematics      & eng\_Latn-tam\_Taml    & 286                                 & Textile             & eng\_Latn-kan\_Knda    & 7630                               \\
Mathematics      & eng\_Latn-mal\_Mlym    & 163                                 & History             & eng\_Latn-tel\_Telu    & 4902                               \\
health           & eng\_Latn-tam\_Taml    & 12728                               & History             & eng\_Latn-mar\_Deva    & 4919                               \\
health           & eng\_Latn-guj\_Gujr    & 299                                 & History             & eng\_Latn-kan\_Knda    & 4941                               \\
health           & eng\_Latn-tel\_Telu    & 34855                               & Economics           & eng\_Latn-tel\_Telu    & 5087                               \\
health           & eng\_Latn-hin\_Deva    & 11802                               & Economics           & eng\_Latn-mar\_Deva    & 5093                               \\
health           & eng\_Latn-mar\_Deva    & 1904                                & Economics           & eng\_Latn-kan\_Knda    & 5061                               \\
health           & eng\_Latn-kan\_Knda    & 135                                 & Tourism             & eng\_Latn-tel\_Telu    & 12798                              \\
health           & eng\_Latn-mal\_Mlym    & 202                                 & Tourism             & eng\_Latn-mar\_Deva    & 12795                              \\
health           & eng\_Latn-ben\_Beng    & 370                                 & Tourism             & eng\_Latn-kan\_Knda    & 12813                              \\
Teaching methods & eng\_Latn-kan\_Knda    & 32670                               & Communication skill & eng\_Latn-tel\_Telu    & 8426                               \\
Teaching methods & eng\_Latn-tel\_Telu    & 31828                               & Communication skill & eng\_Latn-mar\_Deva    & 8537                               \\
Teaching methods & eng\_Latn-hin\_Deva    & 19065                               & Communication skill & eng\_Latn-kan\_Knda    & 8627                               \\
Teaching methods & eng\_Latn-mar\_Deva    & 32327                               &                     &                        & \multicolumn{1}{l}{}              
\end{tabular}%
}
\end{table}

In this section, we introduce the corpora we developed for machine translation between English and 35 Indian languages, covering both general and domain-specific contexts, as detailed in Tables \ref{tab:bhashik-parallel-corpora-generic}, \ref{tab:bhashik-parallel-corpora-domain} and \ref{tab:bhashik-parallel-corpora-domain-health}. These parallel corpora are publicly available for research purposes on Hugging Face \footnote{\url{https://huggingface.co/ltrciiith}} under the names: bhashik-parallel-corpora-generic, bhashik-parallel-corpora-education, and bhashik-parallel-corpora-health.

\section{Automatic Post-Editing (APE) Corpora}

We propose methods to enhance corpora for automatic post-editing between English and Indian languages. These include using human post-edited corpora and generating synthetic corpora from existing parallel data, providing a foundation for general and domain-specific post-editing tasks.

\begin{table}[]
\caption{Automatic post-editing corpora for English and Indian languages: general domain (Bhashik APE Synth - 10M) and education and health domains (Bhashik APE Human - 2M).}
\label{tab:bhashikapecorpus}
\resizebox{\columnwidth}{!}{%
\begin{tabular}{llrlllrllll}
\multicolumn{3}{c}{\textbf{Bhashik APE (Synth) : General}}                                                           & \textbf{} & \multicolumn{3}{c}{\textbf{Bhashik APE (Synth) : General}}                                                           & \textbf{} & \multicolumn{3}{c}{\textbf{Bhashik APE (Human) : Education and Health}}                                \\
\textbf{Source Language} & \textbf{Target Language} & \multicolumn{1}{l}{\textbf{\#Sentences}} &           & \textbf{Source Language} & \textbf{Target Language} & \multicolumn{1}{l}{\textbf{\#Sentences}} &           & \textbf{Source Language} & \textbf{Target Language} & \textbf{\#Sentences}       \\ \hline
eng\_Latn                & san\_Deva                & 167912                                    &           & asm\_Beng                & eng\_Latn                & 307940                                    &           & eng\_Latn                & tam\_Taml                & \multicolumn{1}{r}{13958}  \\
eng\_Latn                & mni\_Mtei                & 75170                                     &           & ben\_Beng                & eng\_Latn                & 498159                                    &           & eng\_Latn                & mar\_Deva                & \multicolumn{1}{r}{733974} \\
eng\_Latn                & ory\_Orya                & 188394                                    &           & brx\_Deva                & eng\_Latn                & 221311                                    &           & eng\_Latn                & kan\_Knda                & \multicolumn{1}{r}{664516} \\
eng\_Latn                & doi\_Deva                & 95643                                     &           & doi\_Deva                & eng\_Latn                & 197047                                    &           & eng\_Latn                & hin\_Deva                & \multicolumn{1}{r}{407093} \\
eng\_Latn                & mai\_Deva                & 124677                                    &           & gom\_Deva                & eng\_Latn                & 201763                                    &           & eng\_Latn                & urd\_Arab                & \multicolumn{1}{r}{4307}   \\
eng\_Latn                & tam\_Taml                & 113849                                    &           & guj\_Gujr                & eng\_Latn                & 327686                                    &           & eng\_Latn                & tel\_Telu                & \multicolumn{1}{r}{751077} \\
eng\_Latn                & mar\_Deva                & 297651                                    &           & hin\_Deva                & eng\_Latn                & 564166                                    &           & eng\_Latn                & mal\_Mlym                & \multicolumn{1}{r}{999}    \\
eng\_Latn                & kan\_Knda                & 135478                                    &           & kan\_Knda                & eng\_Latn                & 432813                                    &           &                          &                          &                            \\
eng\_Latn                & npi\_Deva                & 231413                                    &           & kas\_Arab                & eng\_Latn                & 196620                                    &           &                          &                          &                            \\
eng\_Latn                & hin\_Deva                & 261516                                    &           & kas\_Deva                & eng\_Latn                & 43189                                     &           &                          &                          &                            \\
eng\_Latn                & urd\_Arab                & 225262                                    &           & mai\_Deva                & eng\_Latn                & 156935                                    &           &                          &                          &                            \\
eng\_Latn                & asm\_Beng                & 258234                                    &           & mal\_Mlym                & eng\_Latn                & 524830                                    &           &                          &                          &                            \\
eng\_Latn                & guj\_Gujr                & 146769                                    &           & mar\_Deva                & eng\_Latn                & 274688                                    &           &                          &                          &                            \\
eng\_Latn                & brx\_Deva                & 126492                                    &           & mni\_Beng                & eng\_Latn                & 42669                                     &           &                          &                          &                            \\
eng\_Latn                & ben\_Beng                & 264087                                    &           & mni\_Mtei                & eng\_Latn                & 203056                                    &           &                          &                          &                            \\
eng\_Latn                & gom\_Deva                & 88346                                     &           & npi\_Deva                & eng\_Latn                & 561283                                    &           &                          &                          &                            \\
eng\_Latn                & tel\_Telu                & 192549                                    &           & ory\_Orya                & eng\_Latn                & 385591                                    &           &                          &                          &                            \\
eng\_Latn                & sat\_Olck                & 43932                                     &           & pan\_Guru                & eng\_Latn                & 293203                                    &           &                          &                          &                            \\
eng\_Latn                & kas\_Arab                & 45590                                     &           & san\_Deva                & eng\_Latn                & 217712                                    &           &                          &                          &                            \\
eng\_Latn                & mal\_Mlym                & 723                                       &           & sat\_Olck                & eng\_Latn                & 139586                                    &           &                          &                          &                            \\
tel\_Telu                & eng\_Latn                & 465576                                    &           & snd\_Arab                & eng\_Latn                & 24459                                     &           &                          &                          &                            \\
urd\_Arab                & eng\_Latn                & 457598                                    &           & snd\_Deva                & eng\_Latn                & 93589                                     &           &                          &                          &                            \\
tam\_Taml                & eng\_Latn                & 312192                                    &           &                          &                          & \multicolumn{1}{l}{}                     &           &                          &                          &                           
\end{tabular}%
}
\end{table}

\subsection*{Human APE Corpora}

As discussed in Section \ref{sec:posteditedcorpora}, post-editing involves human translators refining machine-generated translations to produce accurate and natural-sounding final translations. This task was conducted using the PostEditMe\footnote{\url{https://posteditme.in/}} workbench, a platform designed to facilitate post-editing tasks. The workbench allowed translators to edit machine-generated translations while maintaining a record of both the original machine-generated version and the human-corrected version. Additionally, the workbench captured all intermediate stages of post-editing, recording multiple iterations for each sentence pair, along with the final edited version. These intermediate stages and the final post-edited text form a comprehensive resource for studying the post-editing process and its impact on translation quality. The resulting dataset, named the Bhashik APE corpus (presented in Table \ref{tab:bhashikapecorpus}), includes the initial machine-generated (also intermediate stages text) and human-post-edited translations for source texts in English and 11 Indian languages. This corpus supports the development of automatic post-editing systems and enhances domain-specific machine translation by addressing the linguistic nuances of technical and educational contexts more effectively.

\subsection*{Synthetically Developed Corpora}

Automatic post-editing (APE) corpora typically consist of three key components: the source text, an imperfect or erroneous translation, and the error-free translation. To synthetically generate such APE corpora, we utilized both existing parallel corpora and our newly developed parallel corpora described in this work. The target sentences or passages are subjected to various perturbation techniques as detailed in Appendix Section \ref{sec:pertubation}, simulating errors commonly found in machine translations. For our opted language pairs, we processed each sentence pair to create multiple denoised versions of the target text. These perturbations introduced varying degrees of errors, ranging from 2\% to 15\% of the total words, including grammar mistakes, mistranslations, and word order issues. These artificially generated error patterns simulate typical MT errors, making the perturbed version function as the erroneous translation, while the original target text serves as the error-free translation. Additionally, we applied noisy back-translation techniques to create diverse post-editing pairs, ensuring a wide range of error types for robust APE training data. This approach provides a rich and diverse dataset for training automatic post-editing models. We call the resulting dataset Bhashik APE (Synth) and details are presented in Table \ref{tab:bhashikapecorpus}.

\section{Corpora for Machine Translation Evaluation}

In this work, we developed machine translation assessment corpora for English and Indian languages through multiple methods to evaluate translation quality. We focus on translation evaluation which is on a continuous scale from 1 to 100, where 1 indicates poor translation and 100 represents the best possible translation. \\

We use two approaches: reference-based evaluation, comparing translations to a reference, and reference-free evaluation, assessing translations without references. Corpora development involves human expert scoring and synthetic data generation using perturbation techniques (Appendix Section \ref{sec:pertubation}) to simulate common translation errors. Human assessment corpora are created by experts scoring translations on a continuous scale, while synthetic data is generated with COMET models scoring translations across scenarios derived from parallel corpora. This section outlines the strategies for creating synthetic corpora to comprehensively evaluate machine translation systems.
\begin{table}[]
\caption{1M Human direct assessment corpora (without reference) for English and Indian languages across general domain for 36 language pairs.}
\label{tab:humanqeall}
\resizebox{\columnwidth}{!}{%
\begin{tabular}{llr|l|llr}
\textbf{Source Language} & \textbf{Target Language} & \multicolumn{1}{l|}{\textbf{\#Sentences}} &  & \textbf{Source Language} & \textbf{Target Language} & \multicolumn{1}{l}{\textbf{\#Sentences}} \\ \hline
eng\_Latn                & ben\_Beng                & 2960                                      &  & hin\_Deva                & ben\_Beng                & 3001                                     \\
eng\_Latn                & guj\_Gujr                & 2961                                      &  & hin\_Deva                & guj\_Gujr                & 2961                                     \\
eng\_Latn                & hin\_Deva                & 2961                                      &  & hin\_Deva                & kan\_Knda                & 3001                                     \\
eng\_Latn                & kan\_Knda                & 3001                                      &  & hin\_Deva                & mar\_Deva                & 3201                                     \\
eng\_Latn                & mar\_Deva                & 3001                                      &  & hin\_Deva                & ory\_Orya                & 2399                                     \\
eng\_Latn                & ory\_Orya                & 2361                                      &  & hin\_Deva                & pan\_Guru                & 3000                                     \\
eng\_Latn                & Panjabi                  & 2961                                      &  & hin\_Deva                & tel\_Telu                & 3041                                     \\
eng\_Latn                & tam\_Taml                & 2961                                      &  & hin\_Deva                & urd\_Arab                & 2053                                     \\
eng\_Latn                & tel\_Telu                & 2961                                      &  & ben\_Beng                & hin\_Deva                & 3001                                     \\
eng\_Latn                & urd\_Arab                & 1994                                      &  & guj\_Gujr                & hin\_Deva                & 2961                                     \\
ben\_Beng                & eng\_Latn                & 2960                                      &  & kan\_Knda                & hin\_Deva                & 3001                                     \\
guj\_Gujr                & eng\_Latn                & 2961                                      &  & mar\_Deva                & hin\_Deva                & 3201                                     \\
hin\_Deva                & eng\_Latn                & 2961                                      &  & ory\_Orya                & hin\_Deva                & 2080                                     \\
kan\_Knda                & eng\_Latn                & 3001                                      &  & pan\_Guru                & hin\_Deva                & 3000                                     \\
mar\_Deva                & eng\_Latn                & 3001                                      &  & tel\_Telu                & hin\_Deva                & 3041                                     \\
ory\_Orya                & eng\_Latn                & 2321                                      &  & urd\_Arab                & hin\_Deva                & 2001                                     \\
pan\_Guru                & eng\_Latn                & 2961                                      &  & tel\_Telu                & eng\_Latn                & 2961                                     \\
tam\_Taml                & eng\_Latn                & 2961                                      &  & urd\_Arab                & eng\_Latn                & 1961                                    
\end{tabular}%
}
\end{table}
\subsection*{Human Annotated Corpora}

Human-annotated corpora are created by language experts who evaluate translations based on fluency, adequacy, and contextual accuracy, providing a gold standard for assessing MT systems. Experts fluent in both the source and target languages assess the accuracy of the translation in conveying the original meaning, focusing on Translation Accuracy, which determines whether the translation preserves the source text's meaning while maintaining fluency and grammatical correctness. Scores are assigned on a scale of 1 to 100.\\

To develop these corpora, we utilized benchmark datasets and multiple translation engines, including Google Translate, SSMTv3, and Indictransv2, to generate translations for input texts. Experts evaluated both the source text and the translated text, providing ratings. The reference translations in the benchmark corpora were used for reference-enabled machine translation evaluation, while the annotations can also be applied in reference-free evaluation scenarios. The dataset spans 18 language pairs and 36 translation directions, resulting in over 101K rated translations. This effort involved over 70 experienced language experts, with each source and target translation pair receiving three independent ratings. The human-annotated corpora details are presented in Table \ref{tab:humanqeall}.

\begin{table}[]
\caption{10M pseudo direct assessment corpora (with reference) for English and Indian languages across general, educational, and health domains.}
\label{tab:da-all}
\resizebox{\columnwidth}{!}{%
\begin{tabular}{llr|l|llr}
\textbf{Source Language} & \textbf{Target Language} & \multicolumn{1}{l|}{\textbf{\#Sentences}} &  & \textbf{Source Language} & \textbf{Target Language} & \multicolumn{1}{l}{\textbf{\#Sentences}} \\ \hline
asm\_Beng                & eng\_Latn                & 275688                                    &  & eng\_Latn                & hin\_Deva                & 668620                                   \\
ben\_Beng                & eng\_Latn                & 298946                                    &  & eng\_Latn                & mar\_Deva                & 1031650                                  \\
brx\_Deva                & eng\_Latn                & 139431                                    &  & eng\_Latn                & tel\_Telu                & 943632                                   \\
doi\_Deva                & eng\_Latn                & 113633                                    &  & eng\_Latn                & urd\_Arab                & 229569                                   \\
gom\_Deva                & eng\_Latn                & 112863                                    &  & eng\_Latn                & kan\_Knda                & 800024                                   \\
guj\_Gujr                & eng\_Latn                & 154981                                    &  & eng\_Latn                & tam\_Taml                & 127947                                   \\
hin\_Deva                & eng\_Latn                & 356329                                    &  & eng\_Latn                & ben\_Beng                & 264087                                   \\
kan\_Knda                & eng\_Latn                & 354733                                    &  & eng\_Latn                & asm\_Beng                & 258234                                   \\
kas\_Arab                & eng\_Latn                & 138187                                    &  & eng\_Latn                & ory\_Orya                & 188394                                   \\
kas\_Deva                & eng\_Latn                & 42465                                     &  & eng\_Latn                & mai\_Deva                & 124677                                   \\
mai\_Deva                & eng\_Latn                & 148104                                    &  & eng\_Latn                & brx\_Deva                & 126492                                   \\
mal\_Mlym                & eng\_Latn                & 258292                                    &  & eng\_Latn                & doi\_Deva                & 95643                                    \\
mar\_Deva                & eng\_Latn                & 159852                                    &  & eng\_Latn                & npi\_Deva                & 231413                                   \\
mni\_Beng                & eng\_Latn                & 42324                                     &  & eng\_Latn                & guj\_Gujr                & 146769                                   \\
mni\_Mtei                & eng\_Latn                & 121557                                    &  & eng\_Latn                & mni\_Mtei                & 75170                                    \\
npi\_Deva                & eng\_Latn                & 285402                                    &  & eng\_Latn                & san\_Deva                & 167912                                   \\
ory\_Orya                & eng\_Latn                & 206850                                    &  & eng\_Latn                & gom\_Deva                & 88346                                    \\
pan\_Guru                & eng\_Latn                & 37968                                     &  & eng\_Latn                & kas\_Arab                & 45590                                    \\
san\_Deva                & eng\_Latn                & 171657                                    &  & eng\_Latn                & sat\_Olck                & 43932                                    \\
sat\_Olck                & eng\_Latn                & 139451                                    &  & eng\_Latn                & mal\_Mlym                & 1722                                     \\
snd\_Deva                & eng\_Latn                & 62937                                     &  & tel\_Telu                & eng\_Latn                & 331194                                   \\
tam\_Taml                & eng\_Latn                & 129059                                    &  & urd\_Arab                & eng\_Latn                & 255991                                  
\end{tabular}%
}
\end{table}
\begin{table}[]
\caption{70M pseudo direct assessment corpora (without reference) for English and Indian languages across general domain.}
\label{tab:QE-all}
\resizebox{\columnwidth}{!}{%
\begin{tabular}{llr|l|llr}
\textbf{Source Language} & \textbf{Target Language} & \multicolumn{1}{l|}{\textbf{\#Sentences}} &  & \textbf{Source Language} & \textbf{Target Language} & \multicolumn{1}{l}{\textbf{\#Sentences}} \\ \hline
asm\_Beng                & eng\_Latn                & 615562                                    &  & eng\_Latn                & ben\_Beng                & 3050599                                  \\
ben\_Beng                & eng\_Latn                & 101559                                    &  & eng\_Latn                & hin\_Deva                & 4949515                                  \\
bho\_Deva                & eng\_Latn                & 2991                                      &  & eng\_Latn                & guj\_Gujr                & 316079                                   \\
brx\_Deva                & eng\_Latn                & 25297                                     &  & eng\_Latn                & mar\_Deva                & 214706                                   \\
doi\_Deva                & eng\_Latn                & 115211                                    &  & eng\_Latn                & pan\_Guru                & 13973                                    \\
gom\_Deva                & eng\_Latn                & 23858                                     &  & eng\_Latn                & kan\_Knda                & 56693                                    \\
guj\_Gujr                & eng\_Latn                & 307059                                    &  & eng\_Latn                & tam\_Taml                & 10747282                                 \\
hin\_Deva                & eng\_Latn                & 11386159                                  &  & eng\_Latn                & tel\_Telu                & 307867                                   \\
kan\_Knda                & eng\_Latn                & 63748                                     &  & eng\_Latn                & mag\_Deva                & 9292                                     \\
kas\_Arab                & eng\_Latn                & 149232                                    &  & eng\_Latn                & mal\_Mlym                & 8963878                                  \\
kas\_Deva                & eng\_Latn                & 202814                                    &  & eng\_Latn                & npi\_Deva                & 51496                                    \\
lus\_Latn                & eng\_Latn                & 3021                                      &  & eng\_Latn                & kas\_Deva                & 33387                                    \\
mai\_Deva                & eng\_Latn                & 92319                                     &  & eng\_Latn                & brx\_Deva                & 50430                                    \\
mal\_Mlym                & eng\_Latn                & 8883930                                   &  & eng\_Latn                & san\_Deva                & 277518                                   \\
mar\_Deva                & eng\_Latn                & 75266                                     &  & eng\_Latn                & kas\_Arab                & 149251                                   \\
mni\_Beng                & eng\_Latn                & 8775                                      &  & eng\_Latn                & gom\_Deva                & 27831                                    \\
mni\_Mtei                & eng\_Latn                & 25508                                     &  & eng\_Latn                & sat\_Olck                & 25126                                    \\
npi\_Deva                & eng\_Latn                & 1635019                                   &  & eng\_Latn                & mni\_Mtei                & 25471                                    \\
ory\_Orya                & eng\_Latn                & 2959267                                   &  & eng\_Latn                & asm\_Beng                & 109159                                   \\
pan\_Guru                & eng\_Latn                & 27787                                     &  & eng\_Latn                & bho\_Deva                & 2955                                     \\
san\_Deva                & eng\_Latn                & 277532                                    &  & eng\_Latn                & ory\_Orya                & 2936801                                  \\
sat\_Olck                & eng\_Latn                & 25140                                     &  & eng\_Latn                & lus\_Latn                & 3006                                     \\
sin\_Sinh                & eng\_Latn                & 3045                                      &  & eng\_Latn                & mai\_Deva                & 30065                                    \\
snd\_Arab                & eng\_Latn                & 5688                                      &  & eng\_Latn                & sin\_Sinh                & 3012                                     \\
snd\_Deva                & eng\_Latn                & 13114                                     &  & eng\_Latn                & mni\_Beng                & 8855                                     \\
tam\_Taml                & eng\_Latn                & 10614897                                  &  & eng\_Latn                & snd\_Arab                & 5575                                     \\
tel\_Telu                & eng\_Latn                & 351303                                    &  & eng\_Latn                & urd\_Arab                & 46907                                    \\
urd\_Arab                & eng\_Latn                & 47896                                     &  & eng\_Latn                & doi\_Deva                & 24177                                    \\
eng\_Latn                & hoc\_Wara                & 201                                       &  & eng\_Latn                & snd\_Deva                & 13027                                   
\end{tabular}%
}
\end{table}
\subsection*{Synthetically Developed Corpora}

To achieve comprehensive coverage of the translation evaluation spectrum on a linear scale, we recognized that relying solely on existing or limited human scored direct assessment data is insufficient. Therefore, we decided to generate additional data synthetically using existing parallel and benchmark corpora. Various techniques are employed to create degraded, perturbed versions of translations, and we used an existing COMET model to assign degraded scores using the following methodology:

\begin{enumerate}
    \justifying
    \item \textbf{Generating Perturbed Translations:} We began by sourcing parallel and benchmark corpora from above mentioned repositories. For each sentence pair, we applied perturbation techniques outlined in Appendix Section \ref{sec:pertubation} to generate lower-quality translations with varying degrees of errors. The error levels ranged from 5\% to 50\%.
    \item \textbf{Structuring Sentence Tuples:} For each perturbed version, we created a sentence tuple containing the following elements:
    \begin{itemize}
           \item Source text (S)  
           \item Original translation text (T) from the parallel corpora  
           \item Perturbed translation text (\~T) with Error percentage (Ep)  
    \end{itemize}
    \item \textbf{Calculating Synthetic Scores:} Using the existing COMET Quality Estimation (QE) model, we first evaluated the source text (S) and the original translation (T) to obtain a COMET score (Cs) and 100 minus the Translation Error Rate (TER) score (Tp) using original translation (T) and Perturbed translation text (\~T). We then averaged the error percentage (Ep) and the TER-derived score (Tp) to calculate a degradation factor. This factor is subtracted from the COMET score (Cs) to derive a new synthetic score for the source text (S) and the perturbed translation (\~T).

\end{enumerate}

By applying this process, we generated an extensive synthetic dataset to supplement the limited human-annotated data, enabling robust evaluation across a broader range of translation quality scenarios. The resulting dataset is named the Bhashik PseudoMTDA corpus for English and 11 Indian languages. The developed corpora for direct assessment, both with and without reference, are shown in Table \ref{tab:da-all} and Table \ref{tab:QE-all}, respectively.

\section{Corpora for MT Error Span Identification}

\begin{table}[]
\caption{20M pseudo error-annotated corpora with 54M error spans for English and Indian languages across general, educational, and health domains.}
\label{tab:errorall}
\resizebox{\columnwidth}{!}{%
\begin{tabular}{llrrr|l|llrrr}
\textbf{Source Language} & \textbf{Target Language} & \multicolumn{1}{l}{\textbf{\#Sentences}} & \multicolumn{1}{l}{\textbf{\#Errors}} & \multicolumn{1}{l|}{\textbf{\#Errors per Sent}} &  & \textbf{Source Language} & \textbf{Target Language} & \multicolumn{1}{l}{\textbf{\#Sentences}} & \multicolumn{1}{l}{\textbf{\#Errors}} & \multicolumn{1}{l}{\textbf{\#Errors per Sent}} \\ \hline
asm\_Beng                & eng\_Latn                & 551376                                   & 1278030                               & 2.32                                            &  & eng\_Latn                & hin\_Deva                & 1337218                                  & 4849894                               & 3.63                                           \\
ben\_Beng                & eng\_Latn                & 597892                                   & 1385958                               & 2.32                                            &  & eng\_Latn                & ben\_Beng                & 528174                                   & 1205310                               & 2.28                                           \\
brx\_Deva                & eng\_Latn                & 278862                                   & 633800                                & 2.27                                            &  & eng\_Latn                & mar\_Deva                & 2063250                                  & 8725642                               & 4.23                                           \\
doi\_Deva                & eng\_Latn                & 227266                                   & 508742                                & 2.24                                            &  & eng\_Latn                & npi\_Deva                & 727670                                   & 926702                                & 1.27                                           \\
gom\_Deva                & eng\_Latn                & 225726                                   & 493676                                & 2.19                                            &  & eng\_Latn                & asm\_Beng                & 516468                                   & 1237170                               & 2.4                                            \\
guj\_Gujr                & eng\_Latn                & 309962                                   & 705820                                & 2.28                                            &  & eng\_Latn                & san\_Deva                & 337358                                   & 1875910                               & 5.56                                           \\
hin\_Deva                & eng\_Latn                & 712636                                   & 1225358                               & 1.72                                            &  & eng\_Latn                & mni\_Mtei                & 150340                                   & 479422                                & 3.19                                           \\
kan\_Knda                & eng\_Latn                & 707896                                   & 1022196                               & 1.44                                            &  & eng\_Latn                & urd\_Arab                & 459138                                   & 816608                                & 1.78                                           \\
kas\_Arab                & eng\_Latn                & 276374                                   & 663646                                & 2.4                                             &  & eng\_Latn                & mai\_Deva                & 249354                                   & 597684                                & 2.4                                            \\
kas\_Deva                & eng\_Latn                & 84930                                    & 251870                                & 2.97                                            &  & eng\_Latn                & brx\_Deva                & 252984                                   & 684588                                & 2.71                                           \\
mai\_Deva                & eng\_Latn                & 296208                                   & 665144                                & 2.25                                            &  & eng\_Latn                & tel\_Telu                & 1887252                                  & 6017654                               & 3.19                                           \\
mal\_Mlym                & eng\_Latn                & 516584                                   & 1189960                               & 2.3                                             &  & eng\_Latn                & kas\_Arab                & 141542                                   & 244316                                & 1.73                                           \\
mni\_Beng                & eng\_Latn                & 84648                                    & 251446                                & 2.97                                            &  & eng\_Latn                & ory\_Orya                & 379626                                   & 869170                                & 2.29                                           \\
mni\_Mtei                & eng\_Latn                & 243114                                   & 539906                                & 2.22                                            &  & eng\_Latn                & tam\_Taml                & 270474                                   & 375414                                & 1.39                                           \\
npi\_Deva                & eng\_Latn                & 570804                                   & 1317958                               & 2.31                                            &  & eng\_Latn                & doi\_Deva                & 191286                                   & 481650                                & 2.52                                           \\
ory\_Orya                & eng\_Latn                & 413792                                   & 942118                                & 2.28                                            &  & eng\_Latn                & guj\_Gujr                & 293538                                   & 620304                                & 2.11                                           \\
pan\_Guru                & eng\_Latn                & 75936                                    & 158936                                & 2.09                                            &  & eng\_Latn                & kan\_Knda                & 1599988                                  & 5099336                               & 3.19                                           \\
san\_Deva                & eng\_Latn                & 343314                                   & 774460                                & 2.26                                            &  & eng\_Latn                & gom\_Deva                & 353456                                   & 296182                                & 0.84                                           \\
sat\_Olck                & eng\_Latn                & 278902                                   & 620176                                & 2.22                                            &  & eng\_Latn                & sat\_Olck                & 87864                                    & 201198                                & 2.29                                           \\
snd\_Deva                & eng\_Latn                & 125874                                   & 277914                                & 2.21                                            &  & eng\_Latn                & mal\_Mlym                & 311942                                   & 3244                                  & 0.01                                           \\
tam\_Taml                & eng\_Latn                & 258118                                   & 584962                                & 2.27                                            &  & tel\_Telu                & eng\_Latn                & 661054                                   & 954970                                & 1.44                                           \\
urd\_Arab                & eng\_Latn                & 511982                                   & 1183818                               & 2.31                                            &  &                          &                          & \multicolumn{1}{l}{}                     & \multicolumn{1}{l}{}                  & \multicolumn{1}{l}{}                          
\end{tabular}%
}
\end{table}

Error identification and categorization help analyze weaknesses in MT systems, enabling targeted improvements. This section outlines our development of corpora to support these tasks. As discussed in Section \ref{sec:posteditedcorpora}, post-editing involves human translators refining machine-generated translations to produce accurate, natural-sounding final translations. This task was carried out using the PostEditMe\footnote{\url{https://posteditme.in/}} workbench, a platform specifically designed to support post-editing tasks. The workbench captures every edit made by translation experts and records this information at the word level over the machine translation output. The edited spans in the translated text, where changes were made, are marked as MT error spans for the corresponding translation. These post-editing efforts provide a rich resource for identifying error spans and evaluating translation quality. The resulting dataset, named the Bhashik PseudoMTError corpus, includes the source text, the initial machine-generated translation, and the annotated error spans for English and multiple Indian languages.

\subsection*{Synthetically Developed MT Error Span Marked Corpora}

As discussed earlier, corpora for machine translation (MT) error span and category marking typically include three key components: the source text, a translation annotated with error spans, and corresponding error categories. To synthetically create such corpora, we leveraged both existing parallel corpora and newly developed parallel corpora introduced in this work. Target sentences or passages were subjected to various perturbation techniques, detailed in Appendix Section \ref{sec:pertubation}, to simulate common machine translation errors across different categories. These categories were aligned with broader MQM categories.\\

For the selected language pairs, we processed each sentence pair to generate multiple denoised versions of the target text. Perturbations introduced errors ranging from 2\% to 15\% of the total words, including grammar mistakes, mistranslations, and word order issues. These artificially generated error patterns mimic typical MT errors, making the perturbed versions serve as erroneous translations. Error spans were annotated based on the introduced errors to generate the perturbations.\\

This approach provides a rich and diverse dataset for training automatic systems to mark MT error spans and categories. We refer to the resulting dataset as the Bhashik PseudoMTError Corpus and details are presented in Table \ref{tab:errorall}.

\section{Corpora for Grammar Correction}
\label{sec:pertubation}
We use text perturbation techniques to generate grammar correction corpora by introducing controlled errors in monolingual data. These include spelling, grammar, word order, and punctuation errors for all 36 languages. The methods are outlined below:

\begin{itemize}
    \justifying
    \item \textbf{Add/Delete/Replace Random Token:} In this, random tokens are added, deleted, or replaced to simulate common noise found in real text. This introduces variability and errors in the text. For example, adding irrelevant filler words or dropping crucial terms.

    \item \textbf{Change Pronoun:} This perturbation introduces pronoun errors by replacing correct pronouns with incorrect ones, disrupting the coherence and referential clarity of the text. For example, replacing "he" with "she" or "they" can create errors critical for evaluating gender-sensitive, discourse-level translation tasks. To implement this, we used available POS-tagged corpora to compile a list of pronouns for each respective language. For a given input text, a random pronoun is identified and replaced with another randomly selected pronoun from the list.

    \item \textbf{Change Prepositions or Postpositions:} By altering prepositions or postpositions, we introduce syntactic inconsistencies and semantic shifts. For example, replacing "on the table" with "under the table" changes the spatial relationship and meaning. To implement this, we used POS-tagged corpora to compile a list of prepositions or postpositions for each language. Similar to the pronoun perturbation, we randomly select a preposition or postposition in the input text and replace it with another from the predefined list. This perturbation is crucial for tasks focusing on syntax and semantic correctness in translation.

    \item \textbf{Change connectives:} This perturbation modifies logical connectives, introducing errors that affect the flow and coherence of the text. For instance, replacing "because" with "although" or "and" with "but" creates inconsistencies in the logical structure of sentences. To implement this, we curated a list of commonly used connectives for POS tagged corpora of each language. A random connective in the input text is selected and substituted with another from the list. This method helps with errors in logical and discourse-level translation tasks.

    \item \textbf{Change Verb Forms:} This task modifies verb forms to incorrect tense, aspect, or modality. For example, replacing “was running” with “run” or “will ran” introduces errors in temporal and modal constructs that the model must learn to correct.

    \item \textbf{Change Lexical Cohesion:} Words or phrases contributing to the overall cohesion of a passage are replaced with less cohesive random alternatives, disrupting the flow and making the sentence appear disjointed. This method helps with errors in logical and discourse-level translation tasks focusing on lexical cohesion.

    \item \textbf{Change Punctuation:} Punctuation errors, such as missing periods, misplaced commas, or incorrect quotation marks, are introduced to mimic common typographical errors. Correcting such errors helps improve text readability and grammaticality.

    \item \textbf{Grammar:}  In this denoising process, selected words are randomly replaced with grammatically inflected forms derived from a trie structure built using language-specific monolingual corpora. This introduces variations in features like gender, number, person, and verb aspects such as tense and modality. The approach is particularly useful for Indian languages with agglutinative properties, enhancing the understanding and modeling of their complex morphological structures to improve language processing and analysis.

    \item \textbf{Masking:} In this task, specific words are masked to evaluate the model's ability to reconstruct missing information. Tokens are randomly replaced with a single random token using a 0.1\% probability. Inspired by BART, our masking is applied at the token level before subword tokenization. This approach is particularly challenging for Dravidian languages, requiring the model to predict correct linguistic inflections during decoding.

    \item \textbf{Spelling:} Typographical errors are introduced into selected words, affecting around 0.1\% of the text. Spelling perturbation involves adding, removing, or substituting random characters, simulating real-world spelling errors. This helps train the model for tasks requiring spelling correction.

    \item \textbf{Word Order:} Indian languages often have flexible word order. To address this, we apply word order perturbation by randomly rearranging groups of up to four words with a 0.1\% probability. This modifies the original sequence while preserving grammaticality, helping models adapt to structural variations in sentences.
\end{itemize}

\section{Languages}
India's linguistic diversity is rooted in the historical relationships and evolution of languages, shaped by their scripts, families, and groups\footnote{\url{https://language.census.gov.in/}}. Understanding these relationships is essential for developing multilingual models and translation systems. To address this, we modeled 36 languages using their native scripts, ensuring accurate representation and processing as follows:

\subsection{Language Script}
Scripts visually represent languages, influencing orthography and phonetics. Shared scripts, like Devanagari\footnote{\url{https://en.wikipedia.org/wiki/Devanagari}} (used by Hindi, Marathi, Maithili, Bhojpuri, etc.), simplify subword tokenization and multilingual representation. Similarly, the Bengali\footnote{\url{https://en.wikipedia.org/wiki/Bengali_alphabet}} script (used by Bengali, Assamese, Manipuri, etc.) facilitates shared processing. However, languages with distinct scripts, such as Tamil\footnote{\url{https://en.wikipedia.org/wiki/Tamil_language}} (Tamil script), Kannada\footnote{\url{https://en.wikipedia.org/?title=Kannada}} (Kannada script), and Malayalam\footnote{\url{https://en.wikipedia.org/wiki/Malayalam}} (Malayalam script), require different subword vocabularies due to their unique orthographic structures. Additionally, languages like Santali\footnote{\url{https://en.wikipedia.org/wiki/Santali_language}} (Ol Chiki script) and Ho\footnote{\url{https://en.wikipedia.org/wiki/Ho_language}} (Warang Citi script) reflect unique cultural identities, necessitating custom handling. To enable interoperable multilingual systems, script-specific features are mapped to unified representations, balancing shared and unique linguistic properties as part of a custom subword-based tokenizer.

\begin{table}[!ht]
\caption{Language Scripts, IDs, and Groupings by Language Family}
\label{tab:languagetags}
\resizebox{\columnwidth}{!}{%
\begin{tabular}{r|lll}
\multicolumn{1}{l|}{\textbf{ID}} & \textbf{Language Code} & \textbf{Language Group} & \textbf{Language Name (script)}           \\ \hline
1                                & asm\_Beng              & Magadhi                 & Assamese (Bengali script)        \\
2                                & awa\_Deva              & CentralIndic            & Awadhi (Devanagari script)       \\
3                                & ben\_Beng              & Magadhi                 & Bengali (Bengali script)         \\
4                                & bho\_Deva              & Magadhi                 & Bhojpuri (Devanagari script)     \\
5                                & bra\_Deva              & CentralIndic            & Braj Bhasha (Devanagari script)  \\
6                                & brx\_Deva              & TibetoBurman            & Bodo (Devanagari script)         \\
7                                & doi\_Deva              & WesternIndic            & Dogri (Devanagari script)        \\
8                                & eng\_Latn              & WestGermanic            & English (Latin script)           \\
9                                & gom\_Deva              & Maharashtri             & Goan Konkani (Devanagari script) \\
10                               & gon\_Deva              & Dravidian               & Gondi (Devanagari script)        \\
11                               & guj\_Gujr              & WesternIndic            & Gujarati (Gujarati script)       \\
12                               & hin\_Deva              & CentralIndic            & Hindi (Devanagari script)        \\
13                               & hingh\_Deva            & CentralIndic            & Hinglish (Devanagari script)     \\
14                               & hoc\_Wara              & AustroAsiatic           & Ho (Warang Citi script)          \\
15                               & kan\_Knda              & Dravidian               & Kannada (Kannada script)         \\
16                               & kas\_Arab              & WesternIndic            & Kashmiri (Arabic script)         \\
17                               & kas\_Deva              & WesternIndic            & Kashmiri (Devanagari script)     \\
18                               & kha\_Latn              & AustroAsiatic           & Khasi (Latin script)             \\
19                               & lus\_Latn              & TibetoBurman            & Mizo (Latin script)              \\
20                               & mag\_Deva              & Magadhi                 & Magahi (Devanagari script)       \\
21                               & mai\_Deva              & Magadhi                 & Maithili (Devanagari script)     \\
22                               & mal\_Mlym              & Dravidian               & Malayalam (Malayalam script)     \\
23                               & mar\_Deva              & Maharashtri             & Marathi (Devanagari script)      \\
24                               & mni\_Beng              & TibetoBurman            & Manipuri (Bengali script)        \\
25                               & mni\_Mtei              & TibetoBurman            & Meitei (Meitei script)           \\
26                               & npi\_Deva              & CentralIndic            & Nepali (Devanagari script)       \\
27                               & ory\_Orya              & Magadhi                 & Odia (Odia script)               \\
28                               & pan\_Guru              & WesternIndic            & Punjabi (Gurmukhi script)        \\
29                               & san\_Deva              & Vedic                   & Sanskrit (Devanagari script)     \\
30                               & sat\_Olck              & AustroAsiatic           & Santali (Ol Chiki script)        \\
31                               & sin\_Sinh              & Maharashtri             & Sinhala (Sinhala script)         \\
32                               & snd\_Arab              & WesternIndic            & Sindhi (Arabic script)           \\
33                               & snd\_Deva              & WesternIndic            & Sindhi (Devanagari script)       \\
34                               & tam\_Taml              & Dravidian               & Tamil (Tamil script)             \\
35                               & tcy\_Knda              & Dravidian               & Tulu (Kannada script)            \\
36                               & tel\_Telu              & Dravidian               & Telugu (Telugu script)           \\
37                               & urd\_Arab              & CentralIndic            & Urdu (Arabic script)             \\
38                               & xnr\_Deva              & CentralIndic            & Kangri (Devanagari script)      
\end{tabular}%
}
\end{table}

\subsection{Language Family and Group}
We utilized language families and geographical proximity information based on linguistic similarities such as grammar, phonology, and syntax to model the mentioned translation and translation-related tasks. Geographical proximity and cultural exchanges create additional linguistic overlaps beyond family and group classifications. For example, Hindi and Marathi share vocabulary due to geographic closeness, while Tamil and Telugu exhibit shared phonetic and grammatical patterns due to their Dravidian roots. For the languages considered, key families and their groups in the Indian subcontinent include:

\begin{itemize}
\justifying
\item \textbf{Indo-European\footnote{\url{https://en.wikipedia.org/wiki/Indo-European_languages}}}: This family includes various groups such as Central Indic, Western Indic, Magadhi, Maharashtri, and Vedic, encompassing languages with shared syntactic and lexical roots. These commonalities enable efficient shared representations in multilingual models.  
    \begin{itemize}
    \justifying
        \item \textbf{Magadhi}: Includes Assamese (Bengali script), Bengali (Bengali script), Bhojpuri (Devanagari script), Magahi (Devanagari script), Maithili (Devanagari script), and Odia (Odia script), characterized by geographic proximity and linguistic similarities.  
        \item \textbf{Western Indic}: Includes Dogri (Devanagari script), Gujarati (Gujarati script), Kashmiri (Arabic and Devanagari scripts), Punjabi (Gurmukhi script), and Sindhi (Arabic and Devanagari scripts), reflecting shared historical and cultural influences.  
        \item \textbf{Central Indic}: Includes Awadhi (Devanagari script), Braj Bhasha (Devanagari script), Hindi (Devanagari script), Hinglish (Devanagari script), Kangri (Devanagari script), Nepali (Devanagari script), and Urdu (Arabic script), sharing grammatical and lexical roots.  
        \item \textbf{Maharashtri}: Includes Marathi (Devanagari script), Sinhala (Sinhala script), and Goan Konkani (Devanagari script), highlighting regional linguistic ties.  
        \item \textbf{Vedic}: Includes Sanskrit (Devanagari script), a classical language foundational to many modern Indo-European languages in India.  
    \end{itemize}
\item \textbf{Dravidian\footnote{\url{https://en.wikipedia.org/wiki/Dravidian_languages}}}: Includes Kannada (Kannada script), Malayalam (Malayalam script), Tamil (Tamil script), Telugu (Telugu script), Tulu (Kannada script), and Gondi (Devanagari script). These languages are marked by agglutinative morphology and shared structural features.  
\item \textbf{Tibeto-Burman\footnote{\url{https://en.wikipedia.org/wiki/Tibeto-Burman_languages}}}: Includes Bodo (Devanagari script), Manipuri (Bengali script), Meitei (Meitei script), and Mizo (Latin script), characterized by structural similarities despite script diversity.  
\item \textbf{Austro-Asiatic\footnote{\url{https://en.wikipedia.org/wiki/Austroasiatic_languages}}}: Includes Khasi (Latin script), Santali (Ol Chiki script), and Ho (Warang Citi script), each with distinct phonetic and grammatical features.  
\item \textbf{West Germanic}: English (Latin script), often serving as a bridge language in multilingual models.  
\end{itemize}

In computational systems, these relationships support transfer learning. Shared features among Indic languages like Hindi, Bengali, and Marathi improve translation quality for low-resource pairs. Similarly, shared morphology in Dravidian languages enhances generalization across Tamil, Telugu, and Kannada. To uniquely identify languages as presented in Table \ref{tab:languagetags}, we used a tuple structure combining the language family, script, and a three-character code. For instance, Hindi is represented as \texttt{CentralIndic+hin\_Deva}, where \texttt{CentralIndic} denotes the family, \texttt{hin} is the language code, and \texttt{Deva} identifies the Devanagari script.

\subsection{Indian Language specific Tokenizer}
We implemented a SentencePieceProcessor-based\footnote{\url{https://github.com/google/sentencepiece}} subword model for 35 Indian subcontinent languages and English, using a vocabulary size of 48,000 tokens. To train the tokenizer, we randomly mixed 10 million parallel corpora from each language, ensuring robust coverage of script-specific subwords. This approach captures the rich morphological and orthographic variations of Indian languages while maintaining compatibility with English. The shared vocabulary, covering 99.99\% of characters across languages, facilitates multilingual learning, improves translation quality for low-resource pairs, and leverages linguistic similarities among Indian languages, all while ensuring computational efficiency.

\section{Domains}

For machine translation and related applications, we focused on education, law, and healthcare as major domains. The education domain spans diverse disciplines, including Natural Science, Mathematics, Law, and Computer Science, reflecting demand in scientific, legal, and technological communication. Additional fields like Teaching Methods, Psychology, Political Science, Management, Tourism, History, Communication Skills, Economics, Marketing, Textiles, and Health highlight the interdisciplinary nature of educational content. General domains further demonstrate the broad applicability of translation tools for educational and administrative purposes, providing a strong foundation for multilingual education and knowledge dissemination.

\section{Modeling}

BhashaVerse goes beyond traditional translation models by incorporating the aforementioned tasks such as error identification, automatic post-editing, and grammar correction, significantly enhancing translation quality. It is designed to handle the complexities of Indian languages with a medium-range parameter encoder-decoder \citep{vaswani2017attention} model. The model is versatile, supporting fine-tuning and adaptability to newer domains and tasks. Optimized for smaller GPUs, BhashaVerse delivers state-of-the-art performance while remaining accessible for systems with limited computational resources.

\subsection{Input and Output structure for Multi Task Learning}
\label{multitaskobj}

In our learning setup, we utilize multi-task multi-lingual corpora to minimize reliance on additional data. For sequence-to-sequence training, the objective is configured to address current and future tasks. Inspired by structure-driven natural language tasks, we use a JSON-based input-output format \citep{bechard2024reducing}, designed to support current applications and adapt to future use cases. Below is the defined JSON structure:

\noindent\textbf{Input JSON:}
\begin{verbatim}
{
	"task": "task and source to target language information",
	"domain": "",
	"input": { input details based on the task }
}
\end{verbatim}

\noindent\textbf{Output JSON:}
\begin{verbatim}
{
	"task": "task and source to target language information",
	"domain": "",
	"output": { output details based on the task }
}
\end{verbatim}

\begin{itemize}
    \item \textbf{task}: Represents the type of task being performed, such as translation or other language-related tasks. It also specifies the source and target language information. For example, it could be a translation task from English to Hindi or a translation post-editing task from English to Hindi, etc. This provides the model with clear task context.

    \item \textbf{domain}: Indicates the domain of the task, such as general, medical, or legal. This key allows the model to adapt its behavior based on domain-specific requirements. In cases where domain information is not available, then the value would be general.

    \item \textbf{input}: A nested JSON object containing the language information and its corresponding input text for processing.

    \item \textbf{output}: A nested JSON object containing the output language information and the processed output text.
\end{itemize}

This JSON structure ensures clarity and modularity in handling inputs and outputs across tasks. By explicitly defining the task and domain in both the input and output, the model can effectively manage multiple tasks, while the standardized format supports scalability and integration into various applications for new tasks. For this work, we define the task, input, and output sample examples for each of our considered tasks as follows.

\subsubsection{Machine Translation}

\begin{verbatim}
Input Json
{
"task": "Translation
        $WestGermanic+eng_Latn#CentralIndic+hin_Deva", 
"domain": "general", 
"input": {"WestGermanic+eng_Latn": "This committee, the whole 
            state, again, where should be the capital?"}, 
}
Output Json
{
"task": "Translation
         $WestGermanic+eng_Latn#CentralIndic+hin_Deva", 
"domain": "general", 
"output": {"CentralIndic+hin_Latn": "ye samiti, poora pradesh, 
            phir kahaan ho raajadhaanee?"}
}

\end{verbatim}

For the task of translation with English-to-Hindi as an example, the input data format is a JSON object specifying the task as translation between the source language (English, WestGermanic+eng\_Latn) and the target language (Hindi, CentralIndic+hin\_Latn). The domain of the text is set as "general," and the input field contains the English sentence, such as "This committee, the whole state, again, where should be the capital?". The corresponding output JSON contains the Hindi translation in the output field, e.g., "ye samiti, poora pradesh, phir kahaan ho raajadhaanee?". This format trains the model to handle multilingual translation tasks, where the encoder processes the source language text and the decoder generates the target language text.

\subsubsection{Grammar Correction}

\begin{verbatim}
Input Json
{
"task": "Correction$Incorrect 
         WestGermanic+eng_Latn#WestGermanic+eng_Latn", 
"domain": "general", 
"input": {"Incorrect 
           WestGermanic+eng_Latn": "This committee, the whole 
           state, 
           again, whrn shuld be the capital?"}, 
}
Output Json
{
"task": "Correction$Incorrect 
         WestGermanic+eng_Latn#WestGermanic+eng_Latn", 
"domain": "general", 
"output": {"Corrected 
            WestGermanic+eng_Latn": "This committee, the whole
            state,
            again, where should be the capital?"}, 
}

\end{verbatim}

In the grammar correction task, the input JSON specifies the task as correcting grammatically incorrect text for a language. For instance, the input contains an English sentence with errors, such as "This committee, the whole state, again, whrn shuld be the capital?", tagged as Incorrect WestGermanic+eng\_Latn. The output JSON includes the corrected text in the output field, e.g., "This committee, the whole state, again, where should be the capital?". This task trains the model to map ungrammatical inputs to corrected outputs, helping to refine text for improved quality and clarity.

\subsubsection{Machine Translation Post-editing}

\begin{verbatim}
Input Json
{
"task": "Translation post editing
	 $WestGermanic+eng_Latn#Dravidian+tel_Latn", 
"domain": "Computer science", 
"input": {"WestGermanic+eng_Latn": "Light contrast contributes
          to this kind of 
          animation and makes it more impressive.", 
          "Dravidian+tel_Latn": "Tēlikapāṭi kāṇṭrāsṭ ī 
          rakamaina 
          yānimēṣanku dōhadam 
          cēstundi mariyu idi marinta ākaṭṭukuṇṭundi."}
}

Output Json
{
"task": "Translation post editing
	 $WestGermanic+eng_Latn#Dravidian+tel_Latn", 
"domain": "Computer science", 
"output": {"post edited Dravidian+tel_Latn": "Laiṭ kāṇṭrāsṭ ī 
            rakamaina yānimēṣan‌ku dōhadam 
            cēstundi mariyu dānini marinta 
            ākaṭṭukunēlā cēstundi."}
}
\end{verbatim}

This task involves refining machine-translated text for improved accuracy and fluency. For example, for a general domain post-editing task, the input JSON includes the source text in English, the machine-generated translation in Telugu, and the task type as Translation post editing. For example, the English input might be "Light contrast contributes to this kind of animation and makes it more impressive." with a machine-generated Telugu translation "Tēlikapāṭi kāṇṭrāsṭ ī rakamaina yānimēṣanku dōhadam cēstundi mariyu idi marinta ākaṭṭuṇṭundi." The output JSON provides the post-edited Telugu translation, e.g., "Laiṭ kāṇṭrāsṭ ī rakamaina yānimēṣan‌ku dōhadam cēstundi mariyu dānini marinta ākaṭṭukunēlā cēstundi." This data trains the model to refine translations, ensuring better alignment with linguistic norms and domain-specific requirements.

\subsubsection{Machine Translation Direct Assessment}
\begin{verbatim}
Input Json
{
"task": "Translation direct assessment
	 $WestGermanic+eng_Latn#Dravidian+tel_Latn", 
"domain": "Computer science", 
"input": {"WestGermanic+eng_Latn": "Light contrast contributes
          to this kind of animation and makes 
          it more impressive.", 
	   "gold Dravidian+tel_Latn": "Laiṭ kāṇṭrāsṭ ī rakamaina 
          yānimēṣan‌ku dōhadam cēstundi mariyu dānini 
          marinta ākaṭṭukunēlā cēstundi."
	   "system Dravidian+tel_Latn": "Tēlikapāṭi kāṇṭrāsṭ ī 
          rakamaina yānimēṣanku dōhadam cēstundi mariyu 
          idi marinta ākaṭṭukuṇṭundi."}
}

Output Json
{
"task": "Translation direct assessment
	 $WestGermanic+eng_Latn#Dravidian+tel_Latn", 
"domain": "Computer science", 
"output": {"direct assessment score out of 100": 78.34184}
}
\end{verbatim}

In this task, the model is trained to evaluate the quality of a translation against a gold-standard reference. For example, the input JSON includes the source English text, the gold-standard Telugu translation, and the system-generated Telugu translation. For example, the input text might be "Light contrast contributes to this kind of animation and makes it more impressive." with a gold translation and a system-generated translation for comparison. The output JSON provides a direct assessment score, such as 78.34, reflecting the quality of the system output relative to the reference. This trains the model to assign quality scores based on reference-based evaluation metrics.

\subsubsection{Machine Translation Quality Estimation}
\begin{verbatim}
Input Json
{
"task": "Translation quality estimation
	 $WestGermanic+eng_Latn#Dravidian+tel_Latn", 
"domain": "Computer science", 
"input": {"WestGermanic+eng_Latn": "Light contrast contributes
            to this kind of animation and makes 
            it more impressive.", 
        "Dravidian+tel_Latn": "Tēlikapāṭi kāṇṭrāsṭ ī rakamaina 
        yānimēṣanku dōhadam cēstundi mariyu 
        idi marinta ākaṭṭukuṇṭundi."}
}

Output Json
{
"task": "Translation quality estimation
	 $WestGermanic+eng_Latn#Dravidian+tel_Latn", 
"domain": "Computer science", 
"output": {"quality estimation score out of 100": 78.34184}
}
\end{verbatim}

This task evaluates translation quality without a gold-standard reference. For example, the input JSON includes the English source text and its Telugu translation, with the task defined as Translation quality estimation and domain as Computer Science. For example, the input might contain the English sentence "Light contrast contributes to this kind of animation and makes it more impressive." and the Telugu translation is "Tēlikapāṭi kāṇṭrāsṭ ī rakamaina yānimēṣanku dōhadam cēstundi mariyu idi marinta ākaṭṭukuṇṭundi." The output JSON provides a quality score, e.g., 78.34. This format supports training for quality estimation tasks in scenarios where reference translations are unavailable.

\subsubsection{Machine Translation Error Identification}
\begin{verbatim}
Input Json
{
"task": "Translation error marking
	 $WestGermanic+eng_Latn#Dravidian+tel_Latn", 
"domain": "Computer science", 
"input": {"WestGermanic+eng_Latn": "Light contrast contributes 
           to this kind of animation 
           and makes it more impressive.", 
	   "Dravidian+tel_Latn": "Tēlikapāṭi kāṇṭrāsṭ ī rakamaina 
           yānimēṣanku dōhadam cēstundi mariyu 
            idi marinta ākaṭṭukuṇṭundi."}
}

Output Json
{
"task": "Translation error marking
	 $WestGermanic+eng_Latn#Dravidian+tel_Latn", 
"domain": "Computer science", 
"output": {"error marked Dravidian+tel_Latn": 
            "<e>tēlikapāṭi</e> kāṇṭrāsṭ 
            ī rakamaina yānimēṣanku <e>dōhadam</e> cēstundi 
            <e>mariyu</e> <e>idi</e> marinta 
            <e>ākaṭṭukuṇṭundi.</e>"}
}
\end{verbatim}

The error identification task identifies and highlights errors in translated text. Again as an example, the input JSON specifies the source English text and its Telugu translation, with the task labeled as Translation error marking for the domain computer science. For example, the input might include "Light contrast contributes to this kind of animation and makes it more impressive." and the Telugu translation "Tēlikapāṭi kāṇṭrāsṭ ī rakamaina yānimēṣanku dōhadam cēstundi mariyu idi marinta ākaṭṭuṇṭundi." The output JSON highlights errors using tags like <e>. For instance, "<E>tēlikapāṭi</e> kāṇṭrāsṭ <e>dōhadam</e> cēstundi <e>marinta</e> ākaṭṭuṇṭundi." This trains the model to detect and mark specific areas needing improvement in translations.

\subsubsection{Machine Translation Error Identification and Correction}
\begin{verbatim}
Input Json
{
"task": "Translation error marking and correction
	 $WestGermanic+eng_Latn#Dravidian+tel_Latn", 
"domain": "Computer science", 
"input": {"WestGermanic+eng_Latn": "Light contrast contributes 
          to this kind of animation and
          makes it more impressive.", 
	   "Dravidian+tel_Latn": "Tēlikapāṭi kāṇṭrāsṭ ī 
          rakamaina yānimēṣanku dōhadam cēstundi 
          mariyu idi marinta ākaṭṭukuṇṭundi."}
}

Output Json
{
"task": "Translation error marking and correction
	 $WestGermanic+eng_Latn#Dravidian+tel_Latn", 
"domain": "Computer science", 
"output": {"error marked Dravidian+tel_Latn": "<e>tēlikapāṭi</e>
            kāṇṭrāsṭ ī rakamaina yānimēṣanku <e>dōhadam</e> 
            cēstundi <e>mariyu</e> <e>idi</e> marinta 
            <e>ākaṭṭukuṇṭundi.</e>",
	   "post editedDravidian+tel_Latn": "Laiṭ kāṇṭrāsṭ ī 
           rakamaina yānimēṣan‌ku dōhadam cēstundi mariyu 
           dānini marinta ākaṭṭukunēlā cēstundi."}
}
\end{verbatim}

This task combines error marking with automatic correction. The input JSON includes the source text, its corresponding translation, and the defined task as Translation error marking and correction. For example, the English text "Light contrast contributes to this kind of animation and makes it more impressive." and the Telugu translation "Tēlikapāṭi kāṇṭrāsṭ ī rakamaina yānimēṣanku dōhadam cēstundi mariyu idi marinta ākaṭṭuṇṭundi." are provided. The output JSON contains both the error-marked translation (e.g., "<e>tēlikapāṭi</e> kāṇṭrāsṭ <e>dōhadam</e> cēstundi <e>marinta</e> ākaṭṭuṇṭundi.") and the corrected text (e.g., "Laiṭ kāṇṭrāsṭ ī rakamaina yānimēṣan‌ku dōhadam cēstundi mariyu dānini marinta ākaṭṭukunēlā cēstundi."). This format trains the model to not only identify errors but also generate refined, corrected translations.

\subsection{Architecture}

We use an encoder-decoder model using the fairseq toolkit\footnote{\url{https://github.com/facebookresearch/fairseq}} for its ability to handle complex language structures in bilingual and multilingual contexts. The encoder encodes input sentences with positional embeddings, capturing key features and context using self-attention. The decoder generates translations word by word, leveraging self-attention to focus on relevant input parts and maintain coherence with previously generated words \citep{vaswani2017attention}. The transformer encoder-decoder models are trained from scratch using the parameter configurations below.

\begin{itemize}
    \item \textbf{Input:} subword tokens.
    \item \textbf{Embedding Size:} 1024.
    \item \textbf{Feedforward Size:} 8192.
    \item \textbf{Layers:} Encoder: 18, Decoder: 18.
    \item \textbf{Attention Heads:} 16.
    \item \textbf{Dropout:} 0.30.
    \item \textbf{Max Word Sequence Length:} 1200.
    \item \textbf{Batch Size:} 1200 tokens.
    \item \textbf{Initial Learning Rate:} 2e-5.
    \item \textbf{Optimizer:} Adam.
    \item \textbf{Label Smoothing:} 0.1.
    \item \textbf{Precision:} 16-bit floating point.
    \item \textbf{Early Stopping:} Triggered if no increase in training loss for 10 epochs.
    \item \textbf{Beam Size:} 15.
\end{itemize}

\section{Training Process}
The model was trained in two stages using 32 A100 GPUs, each with 40 GB of memory. In the first stage, trained on all available corpora (mentioned in above sections) across specified domains and languages for the considered tasks with JSON format. In the second stage, as a continual training, we focused on exclusive training using human-developed corpora from various sources of different tasks, so that the model would produce outputs that are more aligned with human-like language generation.

\section{Evaluation}
In this section, we present the evaluation of machine translation and related tasks performed on our trained model. These tasks include machine translation, grammar correction, post-editing, human direct assessment of translations, quality assessment, error identification, and combined error identification and correction. For machine translation, we relied on existing benchmark corpora such as Flores \citep{goyal2021flores} and IN22 \citep{gala2023indictrans2}, which provide robust datasets for assessing translation performance across diverse language pairs.\\

However, for the other tasks, we could not find suitable benchmark datasets that covered all the language pairs included in this work. Consequently, we utilized our reserved development corpora (1,000 instances per task per language/pair), specifically removed and curated from the training data, for these tasks. Table \ref{tab:overallresult} outlines the sources and sizes of the corpora used for each task, along with the corresponding evaluation metrics applied to assess model performance. This approach ensures a comprehensive evaluation, particularly for less-studied language pairs and specialized tasks. We discuss the average result for each task as follows.\\

\begin{table}[]
\centering
\resizebox{\columnwidth}{!}{%
\begin{tabular}{l|llllr}
\textbf{Task} & \textbf{Benchmark Source} & \textbf{Language Pairs} & \textbf{\begin{tabular}[c]{@{}l@{}}Size \\ (Sentences)\end{tabular}} & \textbf{Metric} & \multicolumn{1}{l}{\textbf{Average Score}} \\ \hline
\multirow{3}{*}{Machine Translation} & \multirow{3}{*}{FLORES + IN22} & \multirow{3}{*}{\begin{tabular}[c]{@{}l@{}}36 * 36 Languages \\ (English and \\ 35 Indian \\ subcontinent Languages)\end{tabular}} & \multirow{3}{*}{\begin{tabular}[c]{@{}l@{}}517638 \\ + 384560\end{tabular}} & BLEU & 25.45 \\
 &  &  &  & CHAF3 & 53.81 \\
 &  &  &  & COMET-22 & 0.8138 \\
 &  &  &  &  & \multicolumn{1}{l}{} \\ \hline
Grammar Correction & Development Corpora & \begin{tabular}[c]{@{}l@{}}22 Schedule\\  Indian Languages\end{tabular} & \multirow{2}{*}{22000} & BLEU & 90.55 \\
 &  &  &  & CHRF & 97.61 \\
 &  &  &  &  & \multicolumn{1}{l}{} \\ \hline
\begin{tabular}[c]{@{}l@{}}Machine Translation \\ Post-editing\end{tabular} & Development Corpora & \begin{tabular}[c]{@{}l@{}}English to Hindi, Kannada, \\ Marathi, Telugu, Bangla\end{tabular} & \multirow{2}{*}{5000} & BLEU & 84.26 \\
 &  &  &  & CHAF3 & 95.47 \\
 &  &  &  &  & \multicolumn{1}{l}{} \\ \hline
\begin{tabular}[c]{@{}l@{}}Machine Translation\\ Direct Assessment\end{tabular} & Development Corpora & \begin{tabular}[c]{@{}l@{}}English to Hindi, Bengali, \\ Tamil, Telugu, Gujarati, \\ Marathi, Kannada, Malayalam, \\ Assamese, Punjabi, and Urdu\end{tabular} & 11000 & \begin{tabular}[c]{@{}l@{}}Spearman\\ Co-relation\end{tabular} & 0.46202 \\ \hline
\begin{tabular}[c]{@{}l@{}}Machine Translation \\ Quality Estimation\end{tabular} & Development Corpora & \begin{tabular}[c]{@{}l@{}}English to Hindi, Bengali, \\ Tamil, Telugu, Gujarati, \\ Marathi, Kannada, Malayalam, \\ Assamese, Punjabi, and Urdu\end{tabular} & 11000 & \begin{tabular}[c]{@{}l@{}}Spearman\\ Co-relation\end{tabular} & 0.47482 \\ \hline
\begin{tabular}[c]{@{}l@{}}Machine Translation \\ Error Identification\end{tabular} & Development Corpora & \begin{tabular}[c]{@{}l@{}}English to Hindi, Kannada, \\ Marathi, Telugu, Bangla\end{tabular} & 5000 & F1 Score & 41.23 \\ \hline
\multirow{3}{*}{\begin{tabular}[c]{@{}l@{}}Machine Translation \\ Error Identification \\ and Correction\end{tabular}} & \multirow{3}{*}{Development Corpora} & \multirow{3}{*}{\begin{tabular}[c]{@{}l@{}}English to Hindi, Kannada, \\ Marathi, Telugu, Bangla\end{tabular}} & \begin{tabular}[c]{@{}l@{}}Error Identification \\ 5000\end{tabular} & F1 Score & 39.23 \\
 &  &  & \multirow{2}{*}{\begin{tabular}[c]{@{}l@{}}Correction\\ 5000\end{tabular}} & BLEU & 84.02 \\
 &  &  &  & CHAF3 & 94.77 \\ \hline
\end{tabular}%
}
\caption{Performance evaluation of the multi-task, multilingual sequence-to-sequence model with 2 billion parameters across diverse tasks. The table presents the benchmark sources, language pairs, dataset sizes, evaluation metrics, and average scores for machine translation, grammar correction, post-editing, quality assessment, and error identification and correction. }
\label{tab:overallresult}
\end{table}

The model is evaluated on machine translation using benchmark corpora such as FLORES and IN22, covering available languages, including English and other Indian subcontinent languages. With a combined dataset size of 902,198 sentences, the model achieved an average BLEU score \citep{papineni2002bleu} of 25.45, CHRF3 score \citep{popovic2015chrf} of 53.81, and COMET-22 score \citep{rei2022comet} of 0.8138. These results reflect the model's capacity to deliver high-quality translations while effectively handling linguistic diversity and complexity.\\

Multi-task learning significantly enhanced the model's performance across related tasks. For grammar correction, on 22,000 grammatically incorrect sentences, the model achieved an average BLEU and CHRF3 scores of 90.55 and 97.61, respectively. This highlights the model's ability to refine grammatical accuracy while benefiting from its translation training. Similarly, post-editing tasks involving English-to-Hindi, Kannada, Marathi, Telugu, and Bangla translations demonstrated high scores, achieving an average BLEU score of 84.26 and a CHRF3 score of 95.47, showing how translation training supports fine-tuning and error reduction in post-editing.\\

Tasks such as direct assessment and quality estimation, evaluated on 11 Indian languages using 11,000 sentences, further underscore the effectiveness of multi-task learning. The Spearman correlation scores of 0.46202 for direct assessment and 0.47482 for quality estimation reveal the model's alignment with human judgments and its ability to predict translation quality, aided by shared learning across tasks.\\

Error identification and correction tasks, evaluated on five language pairs (English-to-Hindi, Kannada, Marathi, Telugu, and Bangla), demonstrated the model's robustness in identifying noisy translation chunks. Error identification achieved an F1 score of 41.23, while the combined error identification and correction task yielded an average BLEU (84.02) and CHRF3 (94.77) scores for corrected outputs.\\

From the numbers, we can argue that the model effectively handles tasks such as machine translation, grammar correction, post-editing, quality assessment, and error identification and correction across multiple Indian languages. The results demonstrate its superior translation performance and ability to generalize across tasks, highlighting the benefits of multi-task learning.

\section{Conclusion}
This work explores the capabilities of a 2-billion-parameter multilingual, multi-task sequence-to-sequence model in addressing the challenges of translating Indian languages. Through multi-task learning, the model excels in tasks such as machine translation, grammar correction, post-editing, quality estimation, and error identification, effectively handling the syntactic, morphological, and script complexities of Indian languages.\\

The study emphasizes the value of curated datasets, domain-specific corpora, and robust evaluation frameworks in overcoming low-resource language challenges. The model demonstrates scalability across 36 × 36 language pairs, supports domain-specific translations, and ensures contextual coherence in discourse-level tasks, showcasing its versatility.\\

This research, created 10 billion parallel corpora along with millions of related task corpora across 36 Indian subcontinent languages and a trained model, shows the potential to bridge linguistic gaps and support multilingual communication in critical areas such as healthcare, education, and governance. Future work will focus on expanding to additional low-resource languages, enhancing discourse-level performance, and refining automatic post-editing systems, paving the way for advancements in machine translation within resource-constrained settings.

\bibliography{compling_style}

\end{document}